\documentclass[10pt,twocolumn,letterpaper]{article}

\usepackage{iccv}
\usepackage{times}
\usepackage{epsfig}
\usepackage{eso-pic}
\usepackage{graphicx}
\usepackage{amsmath}
\usepackage{amssymb}
\usepackage{booktabs}
\usepackage{enumitem}
\usepackage{multirow}
\usepackage{subfig}
\usepackage{makecell}
\usepackage{textcomp}
\usepackage{diagbox}
\usepackage{threeparttable}
\usepackage[dvipsnames]{xcolor}

\usepackage[pagebackref=true,breaklinks=true,letterpaper=true,colorlinks,bookmarks=false]{hyperref}

\iccvfinalcopy % *** Uncomment this line for the final submission

\def\iccvPaperID{5789} % *** Enter the ICCV Paper ID here

\ificcvfinal\fi

\begin{document}

%%%%%%%%% TITLE
\title{Rethinking and Improving Relative Position Encoding for Vision Transformer
}

\author{Kan Wu$^{1,2,*}$, Houwen Peng$^{2,*,\dagger}$, Minghao Chen$^{2}$, Jianlong Fu$^{2}$, Hongyang Chao$^{1}$ \\ $^1$ Sun Yat-sen University \quad $^2$ Microsoft Research Asia}

\maketitle
% Remove page # from the first page of camera-ready.
\ificcvfinal\thispagestyle{empty}\fi
% add page number
\thispagestyle{plain}
\pagestyle{plain}

%%%%%%%%% ABSTRACT
\begin{abstract}
Relative position encoding (RPE) is important for transformer to capture  sequence ordering of input tokens. General efficacy has been proven in natural language processing. However, in computer vision, its efficacy is not well studied and even remains controversial, e.g., whether relative position encoding can work equally well as absolute position? In order to clarify this, we first review existing relative position encoding methods and analyze their pros and cons when applied in vision transformers. We then propose new relative position encoding methods dedicated to 2D images, called image RPE (iRPE). Our methods consider  directional relative distance modeling as well as the interactions between queries and relative position embeddings in self-attention mechanism. The proposed iRPE methods are simple and lightweight. They can be easily plugged into transformer blocks. Experiments demonstrate that solely due to the proposed encoding methods, DeiT~\cite{deit} and DETR~\cite{detr} obtain up to 1.5\% (top-1 Acc) and 1.3\% (mAP) stable improvements over their original versions on ImageNet and COCO respectively, without tuning any extra hyperparameters such as learning rate and weight decay. Our ablation and analysis also yield interesting findings, some of which run counter to previous understanding. Code and models are open-sourced at \href{https://github.com/microsoft/Cream/tree/main/iRPE}{here}. 

\end{abstract}

\newcommand\blfootnote[1]{% 
\begingroup 
\renewcommand\thefootnote{}\footnote{#1}% 
\addtocounter{footnote}{-1}% 
\endgroup 
}

{
	
	\blfootnote{
	 $^*$Equal contributions. Work performed when Kan and Minghao were interns of MSRA. $^\dagger$ Corresponding author: \href{mailto:houwen.peng@microsoft.com}{\color{black}{houwen.peng@microsoft.com}}
	}
}
%%%%%%%%% BODY TEXT
\vspace{-3mm}
\section{Introduction}
\vspace{-0.0mm}

Transformer recently has drawn great attention in computer vision because of its competitive performance and superior capability in capturing long-range dependencies~\cite{detr, ViT, deit, nonlocal}. The core of transformer is self-attention~\cite{Attention}, which is capable of modeling the relationship of tokens in a sequence. Self-attention, however, has an inherent deficiency --- it cannot capture the ordering of input tokens.  %This limits the transformer to model the highly-structured data, such as 2D images. 
%Therefore, the original transformer~\cite{Attention} proposed to inject explicit embeddingss of position information into the model. 
Therefore, incorporating explicit representations of position information is especially important for transformer, since the model is otherwise entirely invariant to sequence ordering, which is undesirable for modeling structured data.

There are mainly two classes of methods to encode positional representations for transformer. One is absolute, while the other is relative. Absolute methods ~\cite{Fixed_Learned_PE,Attention} encode the absolute positions of input tokens from 1 to maximum sequence length. That is, each position has an individual encoding vector. The encoding vector is then combined with the input token to expose positional information to the model.
On the other hand, relative position methods~\cite{shaw,XL-transformer} encode the relative distance between input elements and learn the pairwise relations of tokens. Relative position encoding (RPE) is commonly calculated via a look-up table with learnable parameters interacting with queries and keys in self-attention modules~\cite{shaw}. Such scheme allows the modules to capture very long dependencies between tokens.
Relative position encoding has been verified to be effective in natural language processing~\cite{bert, XL-transformer, xlnet, T5}. However, in computer vision, the efficacy is still unclear. There are few recent works~\cite{ViT, meituan_rel, bot} shedding light on it, but obtaining controversial conclusions in vision transformers. For example, Dosovitskiy \emph{et al.}~\cite{ViT}
observed that the relative position encoding does not bring any gain comparing to the absolute one (please refer to Tab. 8 in~\cite{ViT}). On the contrary, Srinivas \emph{et al.}~\cite{bot} found that relative position encoding can induce an apparent gain, being superior to the absolute one (please refer to Tab. 4 in~\cite{bot}).
%while using absolute position encodings does not provide as much gain as relative.
Moreover, the mostly recent work~\cite{meituan_rel} claims that the relative positional encoding cannot work equally well as the absolute ones (please refer to Tab. 5 in~\cite{meituan_rel}). These works draw different conclusions on the effectiveness of relative position encoding in models, that motivates us to rethink and improve the usage of relative positional encoding in vision transformer.

On the other hand, the original relative position encoding is proposed for language modeling, where the input data is 1D word sequences~\cite{Attention,XL-transformer,shaw}. But for vision tasks, the inputs are usually 2D images or video sequences, where the pixels are highly spatially structured. 
It is unclear that: 
whether the naive extension from 1D to 2D is suitable for vision models; whether the directional information is important in vision tasks? %These points arouse our interests on re-studying relative position encoding as well.
% These points also motivate us to restudying relative position encoding in vision transformers. %for vision tasks.

In this paper, we first review existing relative position encoding methods, and then propose new methods dedicated to 2D images. We make the following contributions.
%Our methods consider  both  directional  relative  distance  modeling and interactions between query and relative position embeddings in self-attention mechanism.

\begin{itemize}
    \setlength{\itemsep}{5pt}
    \setlength{\parsep}{0pt}
    \setlength{\parskip}{0pt}
    
    \item We analyze several key factors in relative position encoding, including the relative direction, the importance of context, the interactions between queries, keys, values and relative position embeddings, and computational cost. %, and the stability of different methods. 
    The analysis presents a comprehensive understanding of relative position encoding, and provides empirical guidelines for new method design.
    \item We introduce an efficient implementation of relative encoding, which reduces the computational cost from the original $\mathcal{O}(n ^ 2d)$ to $\mathcal{O}(nkd)$, where $k \ll n$. %is much smaller than $n$. 
    Such implementation is suitable for high-resolution input images, such as object detection and semantic segmentation, where the token number might be very large.

     \item We propose four new relative position encoding methods, called image RPE (iRPE), dedicated to vision transformers, considering both efficiency and generalizability. The methods are simple and can be easily plugged into self-attention layers. Experiments show that, without adjusting any hyperparameters 
    and settings, the proposed methods can improve DeiT-S~\cite{deit} and DETR-ResNet50~\cite{detr} by 1.5\% (top-1 Acc) and 1.3\% (mAP) over their original models on ImageNet~\cite{imagenet} and COCO~\cite{coco}, respectively.%, without extra hyperparameter tuning, such as learning rate and weight decay. %Our methods reach state-of-the-art on relative position encodings of vision transformer. 
    
    \item We answer previous controversial questions. We empirically demonstrate that relative position encoding can replace the absolute encoding for image classification task. Meanwhile, the absolute encoding is necessary for object detection, where the pixel position is important for object localization.

\end{itemize}

\section{Background}

\subsection{Self-Attention}
Self-attention plays a fundamental role in transformer. It maps a query and a set of key-value pairs to an output.
More specifically, for an input sequence, \emph{e.g.}, the embeddings of words or image patches, ${\bf{x}} = ({\bf{x}}_1, \ldots, {\bf{x}}_n)$ of $n$ elements where ${\bf{x}}_i \in R^{d_x}$, self-attention computes an output sequence ${\bf{z}} = ({\bf{z}}_1, \ldots, {\bf{z}}_n)$ where ${\bf{z}}_i \in \mathbb{R}^{d_z}$.
Each output element ${\bf{z}}_i$ is computed as a weighted sum of  input elements:
%\vspace{-2mm}
\begin{equation}\label{eq:attn}
%\vspace{-2mm}
{\bf{z}}_i = \sum_{j=1}^{n} \alpha_{ij} ({\bf{x}}_j{\bf{W}}^V).
\end{equation}
Each weight coefficient $\alpha_{ij}$ is computed using a softmax:
%\vspace{-2mm}
\begin{equation}\label{eq:softmax}
%\vspace{-1mm}
\alpha_{ij} = \frac{ \exp({e_{ij}}) }{ \sum_{k=1}^{n} \exp({e_{ik}}) },
\end{equation}
where $e_{ij}$ is calculated using a scaled dot-product attention:
%\vspace{-1mm}
\begin{equation}\label{eq:e}
e_{ij} = \frac{({\bf{x}}_i{\bf{W}}^Q)({\bf{x}}_j{\bf{W}}^K)^T}{\sqrt{d_z}}.
\end{equation}
Here, the projections ${\bf{W}}^Q$, ${\bf{W}}^K$, ${\bf{W}}^V \in \mathbb{R}^{d_x \times d_z}$ are parameter matrices, which are unique per layer.

Rather than computing the self-attention once, Multi-head self-attention (MHSA)~\cite{Attention} runs the self-attention multiple times in parallel, \emph{i.e.}, employing $h$ attention heads. The attention head outputs are simply concatenated and linearly transformed into the expected dimensions. 
% Such multi-head mechanism allows the model to jointly attend to information from different representation subspaces~\cite{Attention}. %at different positions %With a single attention head, averaging inhibits this

\subsection{Position Encoding}
%Recurrent models typically do not use explicit position embeddings since they can learn where they are in the sequence through the recurrent hidden state computation. In our setting, the use of position embeddings requires only a simple addition to the input word embeddings which is a negligible overhead.

%Since transformer contains no recurrence and no convolution, in order for the model to make use of the order of the sequence, we need to inject some information about the relative or absolute position of the tokens in the sequence. To this end, we add ``positional encodings'' to the input embeddings at the bottoms of the encoder and decoder stacks. The positional encodings have the same dimension model as the embeddings, so that the two can be summed. There are many choices of positional encodings, learned and fixed [9].

\textbf{Absolute Position Encoding.} Since transformer contains no recurrence and no convolution, in order for the model to make use of the order of the sequence, we need to inject some information about the position of the tokens. 
The original self-attention considers the \emph{absolute position}~\cite{Attention}, and add the absolute positional encodings ${\bf{p}} = ({\bf{p}}_1, \ldots, {\bf{p}}_n)$ to the input token embedding ${\bf{x}}$ as
%\vspace{-2mm}
\begin{equation}
%\vspace{-2mm}
{\bf{x}}_i = {\bf{x}}_i ~\textcolor{blue}{ +~{\bf{p}}_i },
\end{equation}
where the positional encoding ${\bf{p}}_i, {\bf{x}}_i \in \mathcal{R}^d_x$.
There are several choices of absolute positional encodings, such as the fixed encodings by sine and cosine functions with different frequencies and the learnable encodings through training parameters~\cite{Fixed_Learned_PE,Attention}. 
%\textcolor{red}{Note that, the absolute position encoding is injected only at the input block. Residual connections help propagate position information to higher layers. %In contrast, our proposed relation-aware encodings injects position information at each block. 
%Dehghani et al.~\cite{UniTransformer} and Lan et al.~\cite{Albert} observe better performance by further injecting the position information at each block.}

%In our setting, the use of position embeddings requires only a simple addition to the input word embeddings which is a negligible overhead.

\textbf{Relative Position Encoding.} Besides the absolute position of each input element, recent works also consider the pairwise relationships between elements, \emph{i.e.}, \emph{relative position}~\cite{shaw}. 
Relative relation is presumably important for tasks where the relative ordering or distance of the elements matters.
This type of methods encode the relative position between the input elements ${\bf{x}}_i$ and ${\bf{x}}_j$ into vectors ${\bf{p}}^V_{ij}$, ${\bf{p}}^Q_{ij}$, ${\bf{p}}^K_{ij} \in \mathbb{R}^{d_z}$, where $d_z = d_x$.
The encoding vectors are embedded into the self-attention module, which re-formulates Eq.~(\ref{eq:attn}) and Eq.~(\ref{eq:e}) as
%\vspace{-2mm}
\begin{equation}\label{eq:attn2}
%\vspace{-2mm}
{\bf{z}}_i = \sum_{j=1}^{n} \alpha_{ij} ({\bf{x}}_j{\bf{W}}^V \textcolor{blue}{+~{\bf{p}}^V_{ij}} ) ,
\end{equation}

\begin{equation}\label{eq:e2}
%\vspace{-1mm}
    e_{ij} = \frac{({\bf{x}}_i{\bf{W}}^Q \textcolor{blue}{+~ {\bf{p}}^Q_{ij}})({\bf{x}}_j{\bf{W}}^K \textcolor{blue}{+~ {\bf{p}}^K_{ij}} )^T}{\sqrt{d_z}}.
\end{equation}
In this fashion, the pairwise positional relation is learned during transformer training. Such relative position encoding can be either shared across attention heads or not. %The shared one is more parameter-saving, while being able to achieve comparable performance with the non-shared one, as demonstrated in Tab. \ref{table:buckets_sharing} of the experiment in Sec. \ref{subsec:abstudy}.

\section {Method}

In this section, we first review previous relative position encoding methods and analyze their differences. % in term of both 1D and 2D relative positions. 
Then, we propose four new methods dedicated to vision transformer, and their efficient implementation.

\subsection{Previous Relative Position Encoding Methods}

\begin{figure*}[t]
    \vspace{-7mm}
    \begin{minipage}[b]{1.0\textwidth}
    \centering
    \subfloat[][bias mode]{
                \includegraphics[width=7cm]{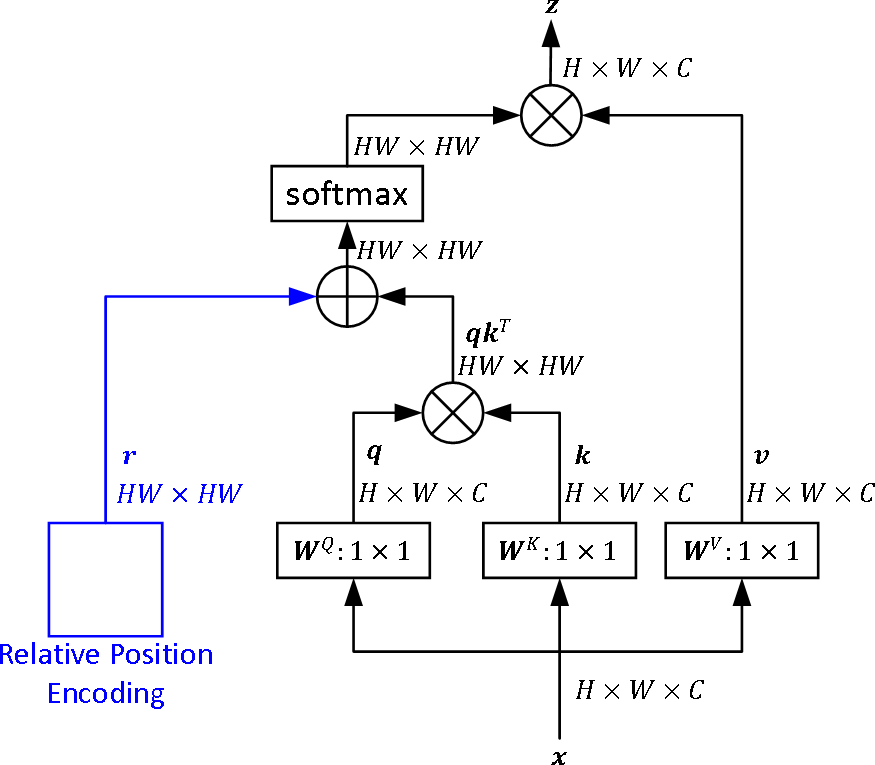}
    }
    \centering
    \subfloat[][contextual mode]{
                \includegraphics[width=7cm]{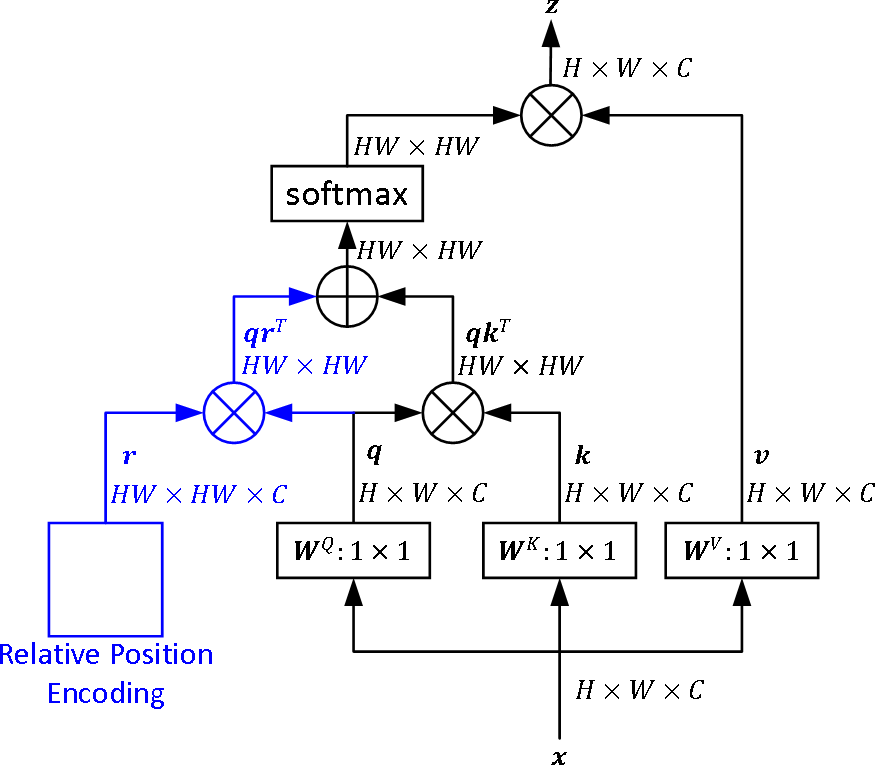}
    }
    \end{minipage}
      \vspace{-3mm}
    \caption {Illustration of self-attention modules with 2D relative position encoding on keys. The \textcolor{blue}{blue} parts are newly added.}
    \label {fig:self_attn_module}
    %\vspace{-3mm}
\end{figure*}

%\textbf{Shaw's Relative Position Encoding.}
\textbf{Shaw's RPE.}
Shaw \emph{et al.}~\cite{shaw} propose a relative position encoding for self-attention. The input tokens are modeled as a directed and fully-connected graph. Each edge between two arbitrary positions $i$ and $j$ is presented by a learnable vector ${\bf{p}}_{ij} \in \mathbb{R}^{d_z}$, namely relative position encoding.  Besides, the authors deemed that precise relative position information is not useful beyond a certain
distance, so introduced a clip function to reduce the number of parameters. The encoding is formulated as 
%\vspace{-2mm}
\begin{equation}\label{eq:shaws_attn2}
%\vspace{-0.5mm}
    {\bf{z}}_i = \sum_{j=1}^{n} \alpha_{ij} ({\bf{x}}_j{\bf{W}}^V \textcolor{blue}{+~{\bf{p}}^V_{clip(i - j, k)}} ),
\end{equation}
\begin{equation}\label{eq:shaws_e2}
    e_{ij} = \frac{({\bf{x}}_i{\bf{W}}^Q)({\bf{x}}_j{\bf{W}}^K \textcolor{blue}{+~ {\bf{p}}^K_{clip(i - j, k)}} )^T}{\sqrt{d_z}},
\vspace{0mm}
\end{equation}
\begin{equation}\label{eq:shaws_clip}
\vspace{0mm}
    clip(x, k) = max(-k, min(k, x)),
\end{equation}
where ${\bf{p}}^V$ and ${\bf{p}}^K$ are the trainable weights of relative position encoding on values and keys,  respectively.
${\bf{p}}^V = ({\bf{p}}_{-k}^V, ..., {\bf{p}}_{k}^V)$ and ${\bf{p}}^K = ({\bf{p}}_{-k}^K, ..., {\bf{p}}_{k}^K)$ where ${\bf{p}}_i^V, {\bf{p}}_i^K \in \mathbb{R}^{d_z}$. The scalar $k$ is the maximum relative distance.

%\textbf{Relative Position Encoding in Transformer-XL.} 
\textbf{RPE in Transformer-XL.} 
Dai \emph{et al.}~\cite{XL-transformer} introduce additional bias terms for queries, and uses the sinusoid formulation for relative position encoding, which is formulated as
%\vspace{-1mm}
\begin{equation}
    e_{ij} = \frac{({\bf{x}}_i{\bf{W}}^Q \textcolor{blue}{+~ {\bf{u}}})({\bf{x}}_j{\bf{W}}^K)^T
    + ({\bf{x}}_i{\bf{W}}^Q\textcolor{blue}{+~ {\bf{v}}}){(\textcolor{blue}{{\bf{s}}_{i - j}{\bf{W}}^R})}^T}{\sqrt{d_z}},
\end{equation}
where ${\bf{u}}, {\bf{v}} \in \mathbb{R}^{d_z}$ are two learnable vectors. 

The sinusoid encoding vector ${\bf{s}}$  provides the prior of relative position~\cite{Attention}.
${\bf{W}}^R \in \mathbb{R}^{d_z \times d_z}$ is a trainable matrix, projecting ${\bf{s}}_{i - j}$ into a location-based key vector.

%\textbf{Huang's Relative Position Encoding.} 
\textbf{Huang's RPE.} 
Huang \emph{et al.}~\cite{better_rpe} propose a new method considering the interactions of queries, keys and relative positions simultaneously. The equation is given as follows
%\vspace{-1mm}
\begin{equation}\label{eq:huang_e2}
%\vspace{-1mm}
    e_{ij} = \frac{({\bf{x}}_i{\bf{W}}^Q \textcolor{blue}{+~ {\bf{p}}_{ij}})({\bf{x}}_j{\bf{W}}^K \textcolor{blue}{+~ {\bf{p}}_{ij}} )^T \textcolor{blue}{-~ {\bf{p}}_{ij}{{\bf{p}}_{ij}}^T }}{\sqrt{d_z}},
\end{equation}
where ${\bf{p}}_{ij} \in \mathbb{R}^{d_z}$ is the relative position encoding shared by queries and keys.

%\textbf{Relative Position Encoding in SASA~\cite{sasa}.} 
\textbf{RPE in SASA.} 
The above three methods are all designed for 1D word sequence in language modeling. Ramachandran \emph{et al.}~\cite{sasa} propose an encoding method for 2D images. The idea is simple. It divides the 2D relative encoding into horizontal and vertical directions, such that each direction can by modeled by a 1D encoding. The method formulation is given as follows
%\vspace{-1mm}
\begin{equation}
%\vspace{-2mm}
    e_{ij} = \frac{({\bf{x}}_i{\bf{W}}^Q)({\bf{x}}_j{\bf{W}}^K \textcolor{blue}{+~ concat({\bf{p}}_{{\delta}\tilde{x}}^K, {\bf{p}}_{{\delta}\tilde{y}}^K)})^T}{\sqrt{d_z}},
\end{equation}
where ${\delta}\tilde{x} = \tilde{x}_i - \tilde{x}_j$ and ${\delta}\tilde{y} = \tilde{y}_i - \tilde{y}_j$ denote the relative position offsets on $x$-axis and $y$-axis of the image coordinate respectively, ${\bf{p}}_{{\delta}\tilde{x}}^K$ and ${\bf{p}}_{{\delta}\tilde{y}}^K$ are learnable vectors with length $\frac{1}{2}d_z$, the $concat$ operation concatenates the two encodings to form a final relative encoding with length of $d_z$. In other words, the same offsets on $x$-axis or $y$-axis share the same relative position encoding, so this method is able to
reduce the number of learnable parameters and computational cost. However, the encoding is only applied on keys. In our experiments, we observe that the RPE imposed on keys, queries and values simultaneously is the most effective one, as presented in Tab.~\ref{table:component_deit} and Tab.~\ref{tab:sota_cls}.
%\textcolor{blue}{We observe that the RPE imposed on keys, queries and values simultaneously is the most effective in Tab.~\ref{table:component_deit},\ref{tab:sota_cls}.}

%\textbf{Relative Position Encoding in Axial-Deeplab \cite{axial_deeplab}.}
\textbf{RPE in Axial-Deeplab.} Wang \emph{et al.} \cite{axial_deeplab} introduce a position-sensitive method that adds qkv-dependent positional bias into self-attention. The position sensitivity is applied on axial attention that propagates information along height-axis and width-axis sequentially. However, when the relative distance is larger than a threshold, the encoding is set to zero. We observe that long-range relative position information is useful, as analysed in Tab.~\ref{table:det_comp}. The position-sensitivity might be competitive when imposed on the standard self-attention. If equipped with the proposed piecewise function, it can be further improved and become more efficient for modeling long-range dependencies.

\subsection{Proposed Relative Position Encoding Methods}

We design our image RPE (iRPE) methods to analyze several factors which are not well studied in prior works (see the analysis in Sec. \ref{subsec:abstudy}). First, to study whether the encoding can be independent of the input embeddings, we introduce two relative position modes: bias and contextual. We present a piecewise function to map relative positions to encodings, being different from the conventional clip function. After that, to study the importance of directivity, we design two undirected and two directed methods. Finally we provide an efficient implementation for our methods.

\textbf{Bias Mode and Contextual Mode.}
Previous relative position encoding methods all depend on input embeddings. It brings a question, \emph{i.e.}, whether the encoding can be independent of the input? We introduce bias mode and contextual mode of relative position encoding to study the question.
The former one is independent of input embeddings, while the latter one considers the interaction with queries, keys or values. More specifically, we introduce a unified formulation as
%\vspace{-1mm}
\begin{equation}\label{eq:eindex}
%\vspace{-1mm}
    e_{ij} = \frac{({\bf{x}}_i{\bf{W}}^Q)({\bf{x}}_j{\bf{W}}^K)^T\textcolor{blue}{ + b_{ij}}}{\sqrt{d_z}},
\end{equation}
where $b_{ij} \in \mathbb{R}$ is the 2D relative position encoding, defining the bias or contextual mode. For bias mode,
%\vspace{-2mm}
\begin{equation}\label{eq:abs_bucket}
%\vspace{-2mm}
    b_{ij} = r_{ij},
\end{equation}
where $r_{ij} \in \mathbb{R}$ is a learnable scalar and represents the  relative position weight between the position $i$ and $j$. For contextual mode,
\begin{equation}\label{eq:ctx_bucket}
%\vspace{-1mm}
    b_{ij} = ({\bf{x}}_i{\bf{W}}^Q){{\bf{r}}_{ij}}^T,
\end{equation}
where ${\bf{r}}_{ij} \in \mathbb{R}^{d_z}$ is a trainable vector, interacted with the query embedding. There are multiple variants for $b_{ij}$ in contextual mode. For example, the relative position encoding operated on both queries and keys can be presented as
\vspace{-0mm}
\begin{equation}\label{eq:ctx_bucket_2}
\vspace{-0mm}
    b_{ij} = ({\bf{x}}_i{\bf{W}}^Q)({{\bf{r}}^K_{ij}})^T + ({\bf{x}}_j{\bf{W}}^K)({{\bf{r}}^Q_{ij}})^T,
\end{equation}
where ${\bf{r}}^K_{ij}, {\bf{r}}^Q_{ij} \in \mathbb{R}^{d_z}$ are both learnable vectors. Besides, contextual mode can also be applied on value embeddings,
% for modeling the pairwise relationship on values by replacing ${\bf{p}}^V_{ij}$ with ${\bf{r}}^V_{ij}$ as 
%\vspace{-2mm}
\begin{equation}\label{eq:ctx_bucket_v}
%\vspace{-2mm}
    {\bf{z}}_i = \sum_{j=1}^{n} \alpha_{ij} ({\bf{x}}_j{\bf{W}}^V \textcolor{blue}{ +~ {{\bf{r}}^V_{ij}}}),
\end{equation}
where ${\bf{r}}^V_{ij} \in \mathbb{R}^{d_z}$. The relative position weights ${\bf{r}}^Q_{ij}$, $ {\bf{r}}^K_{ij}$ and ${\bf{r}}^V_{ij}$ can be constructed in the same way. For a unified representation, we use ${{\bf{r}}_{ij}}$ to denote them in bias mode and contextual mode in the following discussion.
Fig.~\ref{fig:self_attn_module} shows the illustration of self-attention modules with 2D relative position encoding on keys in the propsoed two modes.

\textbf{A Piecewise Index Function.} Before describing the 2D relative position weight ${\bf{r}}_{ij}$, we first introduce a many-to-one function, mapping a relative distance into an integer in finite set, then ${\bf{r}}_{ij}$ can be indexed by the integer and share encondings among different relation positions. Such index function can largely reduce computation costs and the number of parameters for long sequence (\emph{e.g.}, high resolution images). Although the clip function $h(x) = max(-\beta, min(\beta, x))$ used in~\cite{shaw} also reduces the cost, the positions whose relative distance is larger than $\beta$ are assigned to the same encoding. This method inevitably drops out the contextual information of long-range relative positions. Inspired by~\cite{T5}, we introduce a piecewise function $g(x): \mathbb{R} \rightarrow \{y \in \mathbb{Z} | -\beta \le y \le \beta \}$ for indexing relative distances to corresponding encodings. %for real number input and more flexible adjustment of the piecewise point. 
The function is based on a hypothesis that the closer neighbors are more important than the further ones, and distributes the attention by the relative distance. It is presented as
\vspace{-2mm}
\begin{align}
    \begin{split}
    \footnotesize
    g(x) = \left \{
        \begin{array}{lr}
            [x], & |x| \le \alpha \\
            sign(x)\times min(\beta, [\alpha + \frac{\ln{({|x|}/{\alpha}})}{\ln{({\gamma}/{\alpha})}}(\beta - \alpha)]), & |x| > \alpha
        \end{array}
    \right.
    \end{split}
    \label{eq:piecewise_func}
\end{align}
where $[\cdot]$ is a round operation,  $sign()$ determines the sign of a number, \emph{i.e.}, returning 1 for positive input, -1 for negative, and 0 for otherwise. $\alpha$ determines the piecewise point, $\beta$ controls the output in the range of $[-\beta, \beta]$, and $\gamma$ adjusts the curvature of the logarithmic part. 

\begin{figure}[t]

    \centering{
        \includegraphics[width=8.0cm]{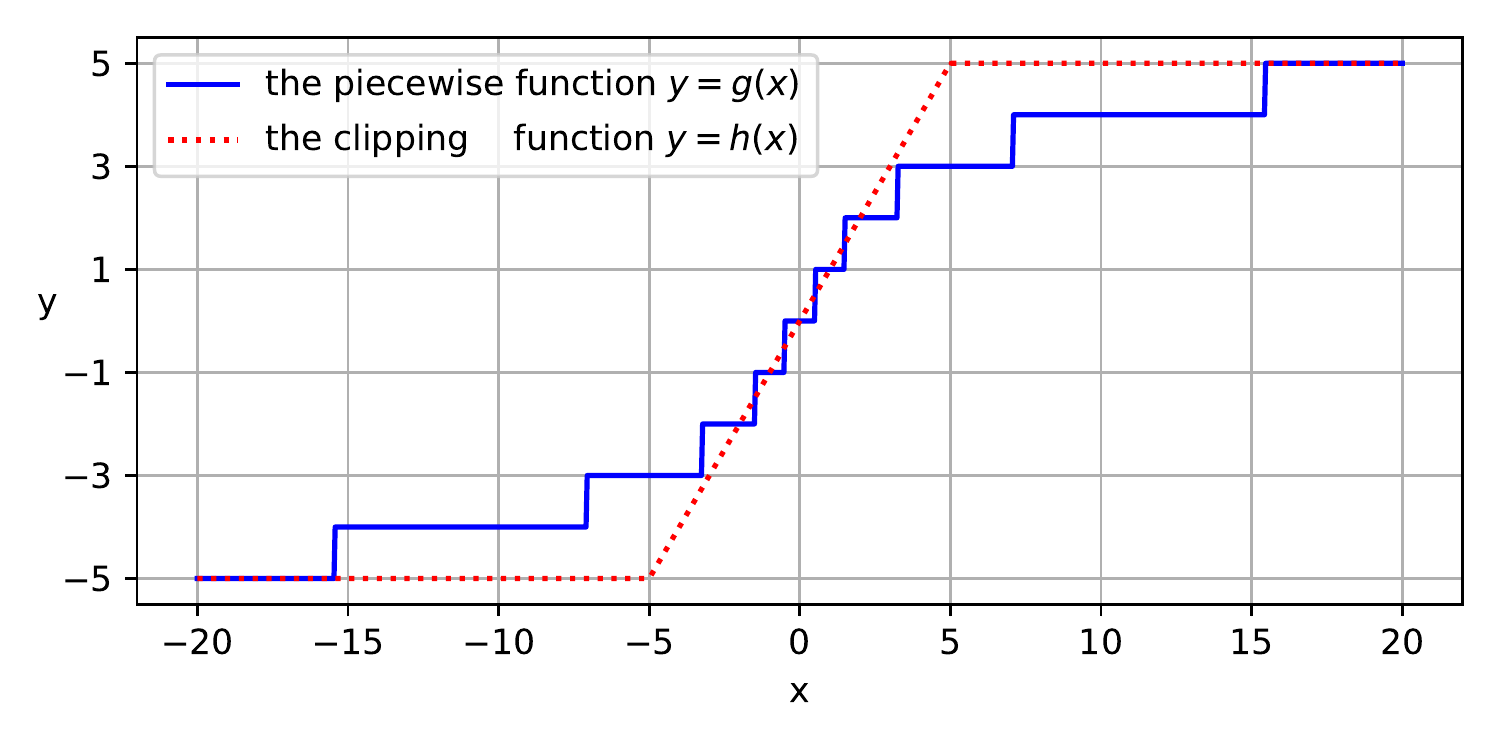}
    }
    \vspace{-6mm}
    \caption {The comparison between the piecewise function $g(x)$ and the clip function $h(x)$.} 
    \label {fig:bucket_map_func}
    %\vspace{-4mm}
\end{figure}

We compare the piecewise function $g(x)$ with the clip function $h(x) = min(-\beta, max(\beta, x))$, \emph{i.e.} Eq.~(\ref{eq:shaws_clip}). In Fig.~\ref{fig:bucket_map_func}, the clip function $h(x)$ distributes uniform attention and leaves out long distance positions, but the piecewise function $g(x)$ distributes different levels of attention by relative distance. We suppose that the potential information in long-range position should be preserved, especially for high resolution images or the tasks requiring long-range feature dependencies, so $g(x)$ is selected to construct our mapping method for ${\bf{r}}_{ij}$.

\textbf{2D Relative Position Calculation.}
In order to calculate relative position on 2D image plane and define the relative weight ${\bf{r}}_{ij}$, we propose two undirected mapping methods, namely Euclidean and Quantization, as well as two directed mapping methods, namely Cross and Product.

\textit{Euclidean method.} On image plane, the relative position $(\tilde{x}_i - \tilde{x}_j, \tilde{y}_i - \tilde{y}_j)$ is a 2D coordinate. We compute Euclidean distance between two positions, and 
maps the distance into the corresponding encoding. The method is undirected and formulated as
%\vspace{-2mm}
\begin{equation}\label{eq:euc_r}
%\vspace{-1mm}
    {\bf{r}}_{ij} = {{\bf{p}}_{I(i, j)}},
\end{equation}
\begin{equation}
%\vspace{-2mm}
     I(i, j)  = g(\sqrt{({\tilde{x}}_i - {\tilde{x}}_j)^2 + ({\tilde{y}}_i - {\tilde{y}}_j)^2}~),   
\end{equation}
where ${{\bf{p}}_{I(i, j)}}$ is either a learnable scalar in bias mode or a  vector in contextual mode. We regard ${{\bf{p}}_{I(i, j)}}$ as a bucket, which stores the relative position weight. The number of buckets is $2\beta + 1$, as defined in Eq.~(\ref{eq:piecewise_func}).

\textit{Quantization method.} In the above Euclidean method, the closer two neighbors with different relative distances may be mapped into the same index, e.g. the 2D relative positions (1, 0) and (1, 1) are both mapped into the index 1. We suppose that the close neighbors should be separated. Therefore, we quantize Euclidean distance, i.e., different real number is mapped into different integer. We revise $I(i, j)$ in Eq.~(\ref{eq:euc_r}) as
%\vspace{-2mm}
\begin{equation}\label{eq:method_quant}
%\vspace{-2mm}
    I(i, j) = g(quant(\sqrt{({\tilde{x}}_i - {\tilde{x}}_j)^2 + ({\tilde{y}}_i - {\tilde{y}}_j)^2})~).
\end{equation}
The operation $quant(\cdot)$ maps a set of real numbers \{0, 1, 1.41, 2, 2.24, ...\} into a set of integers \{0, 1, 2, 3, 4, ...\}. This method is also undirected.

\textit{Cross method.} %The previous two methods are undirected, where encodings are not related to the relative directions. 
%To evaluate the importance of directivity, we propose two directed mapping methods. 
Positional direction of pixels is also important for images, we thereby propose directed mapping methods. 
This method is called Cross method, which computes encoding on horizontal and vertical directions separately, then summarizes them. The method is given as follows,
%\vspace{-2mm}
\begin{equation}
    \label{eq:method_cross}
%\vspace{-1mm}
    {\bf{r}}_{ij} = {\bf{p}}^{\tilde{x}}_{I^{\tilde{x}}(i, j)} + {\bf{p}}^{\tilde{y}}_{I^{\tilde{y}}(i, j)}, 
\end{equation}
\begin{equation}
 \label{eq:method_cross2}
    I^{\tilde{x}}(i, j) = {g({\tilde{x}}_i - {\tilde{x}}_j)},
\vspace{-0.5mm}   
\end{equation}
\begin{equation}
\label{eq:method_cross3}
    I^{\tilde{y}}(i, j) = {g({\tilde{y}}_i - {\tilde{y}}_j)},
\end{equation}
where ${{\bf{p}}^{\tilde{x}}_{I(i, j)}}$ and ${{\bf{p}}^{\tilde{y}}_{I(i, j)}}$ are both learnable scalars in bias mode, or a learnable vectors in contextual mode. Similar to the encoding in SASA~\cite{sasa}, the same offsets on $x$-axis or $y$-axis share the same encoding, but the main difference is that we use a piecewise function to distribute attention by relative distance. The number of buckets is $2 \times (2\beta + 1)$.

\textit{Product method.} %The second directed method is Product. 
The Cross method encodes different relative positions into the same embedding if the distance on one direction is identical, either horizontal or vertical. Besides, the addition operation in Eq.~(\ref{eq:method_cross}) brings extra computational cost. To improve efficiency and involve more directional information, we design Product method which is formulated as below

%\vspace{-2mm}
\begin{equation}\label{eq:method_product}
%\vspace{-2mm}
    {\bf{r}}_{ij} = {\bf{p}}_{I^{\tilde{x}}(i, j), I^{\tilde{y}}(i, j)}.
\end{equation}
The right side of the equation is a trainable scalar in bias mode, or a trainable vector in contextual mode. $I^{\tilde{x}}(i, j)$ and $I^{\tilde{y}}(i, j)$ are defined in Eq.~(\ref{eq:method_cross2}) and Eq.~(\ref{eq:method_cross3}), and the combination of them is a 2D index for ${\bf{p}}$. The number of buckets is $(2\beta + 1)^2$.

\textbf{An Efficient Implementation.} For the above proposed methods in contextual mode, there is a common term  $({\bf{x}}_i{\bf{W}}){{\bf{p}}_{I(i, j)}}^T$ when putting Eq.~(\ref{eq:euc_r}), Eq.~(\ref{eq:method_cross}) or Eq.~(\ref{eq:method_product}) into Eq.~(\ref{eq:ctx_bucket}). Let $y_{ij}$ denote the common term as follows,
%\vspace{-2mm}
\begin{equation}
%\vspace{-2mm}
y_{ij} = ({\bf{x}}_i{\bf{W}}){{\bf{p}}_{I(i, j)}}^T.
\end{equation}
It takes time complexity $\mathcal{O}(n^2d)$ to compute all $y_{ij}$, where $n$ and $d$ are the length of the input sequence and the number of feature channels, respectively.
Due to the many-to-one property of $I(i, j)$, the set size $k$ of $I(i, j)$ is usually less than $n$ in vision transformer. Therefore, we provide an efficient implementation as follows,
%\vspace{-2mm}
\begin{equation}\label{eq:imp_zit}
%\vspace{-2mm}
    z_{i, t} = ({\bf{x}}_i{\bf{W}}){{\bf{p}}_{t}}^T, t \in \{I(i, j)| i, j \in [0, n)\},
\end{equation}
\begin{equation}\label{eq:imp_yij_assign}
    {y_{ij} = z_{i, I(i, j)}}.
\end{equation}
It first takes time complexity $\mathcal{O}(nkd)$ to pre-compute all $z_{i, t}$ by Eq.~(\ref{eq:imp_zit}), then assigns $z_{i, t}$ to all $y_{ij}$ by the mapping $t = I(i, j)$ by Eq.~(\ref{eq:imp_yij_assign}). The assignment operation takes time complexity $\mathcal{O}(n^2)$, whose cost is much smaller than that of the pre-computation procedure. Thus, the computational cost of relative position encoding reduces from the original $\mathcal{O}(n^2d)$ to $\mathcal{O}(nkd)$.

\section {Experiments}
\label{sec:exp}
    In this section, we first provide some analysis by comparing different position embeddings, followed by experiments on the effects of key factors in relative position encoding. Then, we compare the proposed methods with the state-of-the-art methods on image classification and object detection tasks. Finally, we visualize the relative position encoding and explain why it works.

\subsection{Implementation Details}
We choose the recent vision transformer model DeiT~\cite{deit} as the baseline for most experiments. The relative position encoding is added into all self-attention layers. If not specified, the relative position encoding is only added on keys. We set $\alpha$:$\beta$:$\gamma$ = 1:2:8 for the piecewise function $g(x)$, and adjust the number of buckets by changing $\beta$. An extra bucket is used to store the relative position encodings of the classification token.

For fair comparison, we adopt the same training settings as DeiT~\cite{deit}: AdamW~\cite{adamw} optimizer with weight decay 0.05, initial learning rate $1$x$10^{-3}$ and minimal learning $1$x$10^{-5}$ with cosine scheduler, 5 epochs warmup, batch size of 1024, 0.1 label smoothing~\cite{labelsmooth}, and stochastic depth with survival rate probability of 0.9. For training, the images are split into 14x14 non-overlapping patches.
Data augmentation methods~\cite{mixup, cutmix} are also consistent with DeiT~\cite{deit}. 
All models are trained from scratch for 300 epochs with 8 NVIDIA Tesla V100 GPUs.

\subsection{Analysis on Relative Position Encoding}
\label{subsec:abstudy}

\begin{table}

\scalebox{0.835}{
    \begin{tabular} {ccccc}
        \Xhline{2\arrayrulewidth}
        Method & \small{Is} & \multirow{2}*{Mode} & Top-1 & $\Delta$ \\
        based on DeiT-S~\cite{deit} & \small{Directed} & ~ & Acc(\%) & Acc(\%) \\
        \Xhline{2\arrayrulewidth}
            Original~\cite{deit} & - & - & 79.9 & -- \\
            % deit_baseline_lab9
        \hline
            \multirow{2}*{Euclidean} & \multirow{2}*{\texttimes} & bias & 80.1 & +0.2 \\ 
            ~ & ~ & contextual & 80.4 & +0.5 \\
            % bias_euc_lab1
            % ctx_euc_lab5 
        \hline
            \multirow{2}*{Quantization} & \multirow{2}*{\texttimes} & bias & 80.3 & +0.4 \\ 
            ~ & ~ & contextual & 80.5 & +0.6 \\
            % bias_quant_51_lab13 
            % ctx_quant_51_lab13
        \hline
            \multirow{2}*{Cross} & \multirow{2}*{\checkmark} & bias & 80.5 & +0.6 \\ 
            ~ & ~ & contextual & 80.8 & +0.9 \\
            % bias_cross_lab9 
            % ctx_cross_lab9
        \hline
            \multirow{2}*{Product} & \multirow{2}*{\checkmark} & bias & 80.5 & +0.6 \\ 
            ~ & ~ & contextual & 80.9 & +1.0 \\
            % bias_product_lab1
            % ctx_product_50_lab7 
        \Xhline{2\arrayrulewidth}
    \end{tabular}
}
    \caption {Ablation of our relative position encoding methods on ImageNet~\cite{imagenet}. The original model is DeiT-S~\cite{deit}, which only uses absolute position encoding. We equip the model with the proposed four relative encoding methods, \emph{i.e.}, Eq.~(\ref{eq:euc_r}), Eq.~(\ref{eq:method_quant}), Eq.~(\ref{eq:method_cross}) and Eq.~(\ref{eq:method_product}). We select the best numbers of buckets for each method, which are 20, 51, 56 and 50 respectively. 
    }
    \label {table:buckets_mapping}
    %\vspace{-3mm}
\end{table}

\textit{Directed \emph{v.s.} Undirected.} 
As shown in Tab. \ref{table:buckets_mapping}, directed methods (Cross and Product), in general, perform better than undirected ones (Euclidean and Quantization) in vision transformer. This phenomenon illustrates that the directivity is important for vision transformers, because image pixels are highly structured and semantically correlative.

\textit{Bias \emph{v.s.} Contextual.}
Tab. \ref{table:buckets_mapping} shows that the contextual mode achieves superior performance to that of bias mode, regardless of which method uses. The underlying reason might be that contextual mode changes the encoding with the input feature while bias mode keeps static.

\textit{Shared \emph{v.s.} Unshared.}
Self-attention contains multiple heads. The relative position encoding can be either shared or unshared across different heads.
We show the effects of these two schemes in bias and contextual modes in Tab. \ref{table:buckets_sharing}, respectively. 
For bias mode, the accuracy drops significantly when sharing encoding across the heads. By contrast, in contextual mode, the performance gap between two schemes is negligible. Both of them achieve an average top-1 accuracy of 80.9\%. We conjecture that different head needs different relative position encoding (RPE for short) to capture different information. In contextual mode, each head can compute its own RPE by the Eq.~(\ref{eq:ctx_bucket}) while in bias mode the shared RPE forces all heads to pay the same attention on patches. For parameter-saving, we adopt the share scheme in our final methods.

\begin{table}[t]
    \centering
    \scalebox{0.95}{
    \begin{tabular}{c|c|cc|c}
        \Xhline{2\arrayrulewidth}
        \multirow{2}*{Mode} & \multirow{2}*{Shared} & {\#Param.} & {MACs} & {Top-1} \\
        ~ & ~ & {(M)} & {(M)} & {Acc(\%)} \\
        \Xhline{2\arrayrulewidth}
                \multirow{2}*{Bias} & \texttimes & 22.05 & 4613 & $80.54 \pm 0.06$ \\
                ~ & \checkmark & 22.05 & 4613 & $80.05 \pm 0.04$ \\
        \hline
                \multirow{2}*{Contextual} & \texttimes & 22.28 & 4659 & $80.99 \pm 0.16$ \\
                % ctx_product_50_lab7
                % ctx_product_50_lab5
                % ctx_product_lab6
                ~ & \checkmark & 22.09 & 4659 & $80.89 \pm 0.04$ \\
                % ctx_product_shared_50_lab3
                % ctx_product_shared_50_lab5
                % ctx_product_shared_50_lab16
        \Xhline{2\arrayrulewidth}
    \end{tabular}
    }
    \caption{Ablation of shared and unshared relative position encoding across attention heads. The experiments are conducted over DeiT-S~\cite{deit} on ImageNet~\cite{imagenet} with 50 buckets. The models are trained and evaluated by three times.}
    \label{table:buckets_sharing}
    %\vspace{-3mm}
\end{table}

\begin{table}
    \scalebox{0.952}{
        \begin{tabular}{c|c|c|c}
            \Xhline{2\arrayrulewidth}
            Function & Mode & Top-1 Acc(\%) & Top-5 Acc(\%) \\
            \Xhline{2\arrayrulewidth}
            \multirow{2}*{clip} & bias & 80.1 & 94.9 \\  % bias_clip_shared_50_lab20
            ~ & contextual & 80.9 & 95.5 \\ % ctx_clip_shared_50_lab20
            \hline
            \multirow{2}*{piecewise} & bias & 80.0 & 95.0 \\
            ~ & contextual & 80.9 & 95.5 \\
            \Xhline{2\arrayrulewidth}
        \end{tabular}
    }
    \caption{Ablation for clip function and piecewise function. The experiments are conducted over DeiT-S~\cite{deit} model with product shared-head relative position encoding on ImageNet~\cite{imagenet}. The number of buckets is 50. }
    \label{tab:function_cmp}
    %\vspace{-3mm}
\end{table}

\textit{Piecewise \emph{v.s.} Clip.} We compare the efficacy of the piecewise function $g(x)$ defined in Eq.~(\ref{eq:piecewise_func}) and the clip function $h(x)$ defined in Eq.~(\ref{eq:shaws_clip}) in Tab. \ref{tab:function_cmp}. There is a very small, even negligible, performance gap between these two functions in image classification task. However, in object detection task, we found that clip function is worse than the piecewise one as illustrated in Tab. \ref{table:det_comp} (\#5 \emph{v.s.} \#6).
% We suppose that the difference lies in the resolution.
The underlying reason is that the two functions are very similar when the sequence length is short. The piecewise function is effective especially when the sequence size is much larger than the number of buckets. Object detection uses a much higher resolution input compared to classification, leading to a much longer input sequence. We therefore conjecture that when the input sequence is long, the piecewise function should be used since it is able to distribute different attentions to the positions with relative large distance, while the clip function assigns the same encoding when the relative distance is larger than $\beta$. 

\begin{figure}[t]
    \centerline{\includegraphics[width=8.0cm]{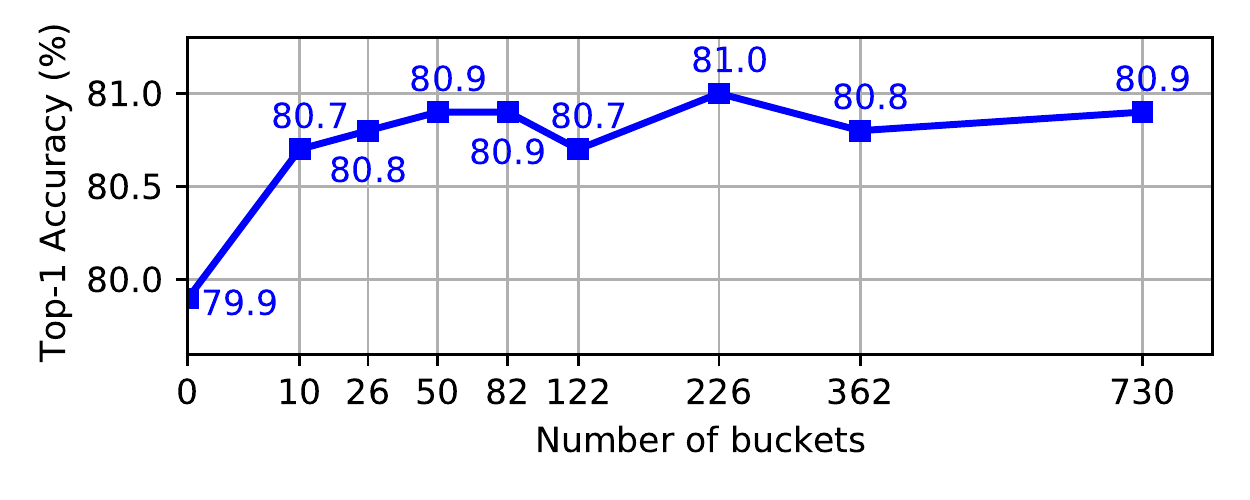}}
    \vspace{-4mm}
    \caption {Ablation for the number of buckets in contextual product model with shared relative position encodings on ImageNet~\cite{imagenet}.}
    \label {fig:bucket_num}
\end{figure}
%\vspace{-1mm}

\begin{figure}[t]
%\vspace{-1mm}
    \centerline{\includegraphics[width=8.0cm]{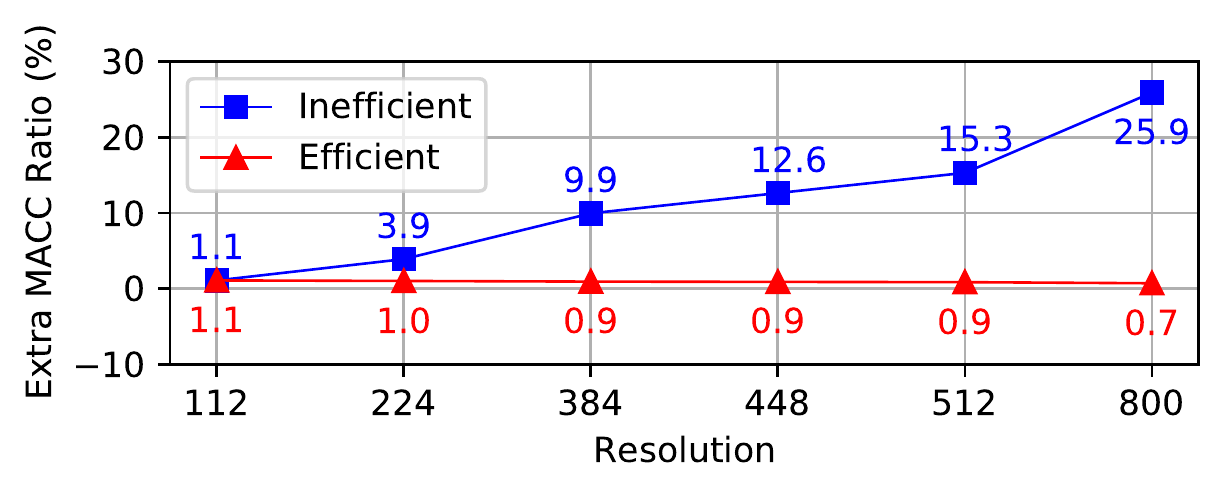}}
    \vspace{-2mm}
    \caption {The extra computational cost of relative position encoding with different implementation in different resolutions. The baseline model is DeiT-S~\cite{deit}. The number of buckets is 50. MACs means multiply-accumulate operations.}
    %\vspace{-3mm}
    \label {fig:compute_cost}
\end{figure}

\textit{Number of buckets.} 
The number of buckets largely affects model parameters, computational complexities and performance. 
In order to find a balance, we explore the influence of varying the number of buckets for the contextual Product method. Fig.~\ref{fig:bucket_num} shows the change of top-1 accuracy along with the number of buckets. The accuracy increase from 79.9 to 80.9 before 50 buckets. After that, there is no significant improvement. It shows that the number of buckets 50 is a good balance between the computational cost and the accuracy for $14 \times 14$ feature map in DeiT-S~\cite{deit}.

\begin{table}[t]
    \centering
    \begin{tabular}{c|cccc|cc}
    \Xhline{2\arrayrulewidth}
        \# & Abs Pos. & $p_{ij}^Q$ & $p_{ij}^K$ & $p_{ij}^V$ & Top-1 & Top-5 \\
    \Xhline{2\arrayrulewidth}
        1~\cite{deit} & learnable & \texttimes & \texttimes & \texttimes & 79.9 & 95.0 \\ % deit_baseline_lab9 
    \hline
        2 & \texttimes & \texttimes & \texttimes & \texttimes & 77.6{\textcolor{red}{\scriptsize{(-2.3)}}} & 93.8 \\ % baseline_noabspos_lab12
        3 & \texttimes & \checkmark & \texttimes & \texttimes & 80.9{\textcolor{YellowGreen}{\scriptsize{(+1.0)}}} & 95.4 \\ % ctx_product_shared_only_kr_trans_noabs_lab26
        4 & \texttimes & \texttimes & \checkmark & \texttimes & 80.9{\textcolor{YellowGreen}{\scriptsize{(+1.0)}}} & 95.3 \\ %ctx_product_shared_noabs_lab18 
        5 & \texttimes & \texttimes & \texttimes & \checkmark & 80.2{\textcolor{YellowGreen}{\scriptsize{(+0.3)}}} & 95.0 \\ % ctx_product_shared_v_noabs_lab18
        6 & \texttimes & \checkmark & \checkmark & \texttimes & 81.0{\textcolor{YellowGreen}{\scriptsize{(+1.1)}}} & 95.5 \\ % ctx_product_shared_qkrv_trans_noabs_lab26 
        7 & \texttimes & \checkmark & \checkmark & \checkmark & 81.3{\textcolor{YellowGreen}{\scriptsize{(+1.4)}}} & \textbf{95.7} \\ % ctx_product_shared_qkrv_trans_noabs_lab26
        8 & learnable & \checkmark & \texttimes & \texttimes & 80.9{\textcolor{YellowGreen}{\scriptsize{(+1.0)}}} & 95.5 \\ % ctx_product_shared_only_kr_trans_lab26
        9 & learnable & \texttimes & \checkmark & \texttimes & 80.9{\textcolor{YellowGreen}{\scriptsize{(+1.0)}}} & 95.5 \\ % ctx_product_shared50_lab3  
        10 & learnable & \texttimes & \texttimes & \checkmark & 80.2{\textcolor{YellowGreen}{\scriptsize{(+0.3)}}} & 95.1 \\ % ctx_product_shared_v_lab18
        11 & learnable & \checkmark & \checkmark & \texttimes & 81.1{\textcolor{YellowGreen}{\scriptsize{(+1.2)}}} & 95.4 \\ % ctx_product_shared_qkr_trans_lab26
        12 & learnable & \checkmark & \checkmark & \checkmark & \textbf{81.4{\textcolor{YellowGreen}{\scriptsize{(+1.5)}}}} & 95.6 \\ % ctx_product_shared_qkrv_trans_lab26
    \Xhline{2\arrayrulewidth}
\end{tabular}

    \caption {Component-wise analysis on ImageNet~\cite{imagenet}. We add contextual product shared-head relative position encodings into DeiT-S~\cite{deit}. The number of buckets is 50. Abs Pos. represents the absolute position encoding. $p_{ij}^Q$, $p_{ij}^K$ and $p_{ij}^V$ present relative position encodings on queries, keys and values.}
    \label {table:component_deit}
    %\vspace{-3mm}
\end{table}

\textit{Component-wise analysis.} We perform a component-wise analysis to study the effects of different position encodings for vision transformer models. We select DeiT-S model~\cite{deit} as the baseline, and only change the position encoding methods. The learnable absolute position encoding is used in the original model.
The relative position encodings are computed by contextual Product method with 50 buckets. The conclusions we got from Tab. \ref{table:component_deit} are as follows: 1) Removing absolute position encoding from original DeiT-S will cause that the Top-1 accuracy drops from 79.9 to 77.6 (\#1 \emph{v.s.} \#2).
2) The models with only relative position encoding surpass the one with only absolute position encoding (\#3-5 \emph{v.s.} \#1). It shows that relative position encoding works well as the absolute one.
3) When equipped with relative position encoding, the absolute one does not bring any gains (\#3-5 \emph{v.s.} \#8-10). We suppose that the local information is more important than the global one in classification task.
4) The relative position encoding on queries or keys brings more gain than that on values (\#3,4 \emph{v.s.} \#5).
5) The combination of the encodings on queries, keys and values brings further improvements (\#6,7,11,12 \emph{v.s.} others).

\textit{Complexity Analysis.} We evaluate the computational cost of our proposed methods with respect to different input resolutions. The baseline model is DeiT-S~\cite{deit} with only absolute position encoding. We adopt contextual product shared-head relative position encoding to the baseline with 50 buckets. Fig.~\ref{fig:compute_cost} shows our method takes at most 1\% extra computational cost with efficient implementation.

\begin{table}[t]

    \begin{threeparttable}
\scalebox{0.84}{
\begin{tabular}{c|c|cc|c}
    \Xhline{2\arrayrulewidth}
    \multirow{2}*{Model} & \multirow{2}*{\#Param.} & \multirow{2}*{Input} & MACs & Top-1 \\
    ~ & ~ & ~ & (M) & Acc (\%) \\
    \Xhline{2\arrayrulewidth}
    \multicolumn{5}{c}{Convnets} \\
    \hline
    ResNet-50~\cite{resnet} & 25M & $224^2$ & 4121 & 79.0 \\
    RegNetY-4.0GF~\cite{radosavovic2020designing} & 21M & $224^2$ & 4012 & 79.4 \\

    EfficientNet-B1~\cite{efficientnet} & 8M & $240^2$ & 712 & 79.1 \\
    EfficientNet-B5~\cite{efficientnet} & 30M & $456^2$ &  10392 & 83.6 \\
    % EfficientNet-B6~\cite{efficientnet} & 43M & $528^2$ &  19297 & 84.0 \\
    % EfficientNet-B7~\cite{efficientnet} & 66M & $600^2$ &  38195 & 84.3 \\
    \hline
    \multicolumn{5}{c}{Transformers} \\
    \hline
    ViT-B/16~\cite{ViT} & 86M & $384^2$ & 55630 & 77.9 \\
    ViT-L/16~\cite{ViT} & 307M & $384^2$ & 191452 & 76.5 \\
    \hline
    DeiT-Ti~\cite{deit} & 5M & $224^2$ & 1261 & 72.2 \\
    CPVT-Ti(0-5)~\cite{meituan_rel} & 6M & $224^2$ & 1262 & 73.4 \\
    DeiT-Ti with {iRPE-K}\textbf{(Ours)} & 6M & $224^2$ & 1284 & 73.7 \\
    \hline
    DeiT-S~\cite{deit} & 22M & $224^2$ & 4613 & 79.9 \\
    CPVT-S(0-5)~\cite{meituan_rel} & 23M & $224^2$ & 4616 & 80.5 \\
    DeiT-S\footnotesize{(Shaw's)}~\cite{deit, shaw}$^+$ & 22M & $224^2$ & 4659 & 80.9\\
    DeiT-S\footnotesize{(Trans.-XL's)}~\cite{deit, XL-transformer}$^+$ & 23M & $224^2$ & 4828 & 80.8 \\
    DeiT-S\footnotesize{(Huang's)}~\cite{deit, better_rpe}$^+$ & 22M & $224^2$ & 4706 & 81.0 \\
    DeiT-S\footnotesize{(SASA's)}~\cite{deit, sasa}$^{*}$ & 22M & $224^2$ & 4639 & 80.8 \\
    DeiT-S with {iRPE-K}\textbf{(Ours)} & 22M & $224^2$ & 4659 & 80.9 \\
    DeiT-S with {iRPE-QK}\textbf{(Ours)} & 22M & $224^2$ & 4706 & 81.1 \\
    DeiT-S with {iRPE-QKV}\textbf{(Ours)} & 22M & $224^2$ & 4885 & 81.4 \\
    \hline
    DeiT-B~\cite{deit} & 86M & $224^2$ & 17592 & 81.8 \\
    CPVT-B(0-5)~\cite{meituan_rel} & 86M & $224^2$ & 17598 & 81.9 \\
    DeiT-B with {iRPE-K}\textbf{(Ours)} & 87M & $224^2$ &  17684 & 82.4 \\
    \Xhline{2\arrayrulewidth}
    \end{tabular}
}
        \begin{tablenotes}
            \footnotesize
            \item [+] We utilize our product method to adapt 1D encoding for 2D images with \\ 
                    the clip function. The encoding weight is shared across heads.
            \item [*] DeiT-S~\cite{deit} with SASA~\cite{sasa}'s relative position encoding.
        \end{tablenotes}
    \end{threeparttable}
    \caption {Comparison on ImageNet~\cite{imagenet}.}
    %\vspace{-3mm}
    \label {tab:sota_cls}
\end{table}

\begin{table*}[t]
\centering
\begin{tabular}{c|cc|c|c|c|cccccc}
    \Xhline{2\arrayrulewidth}
    \# & Abs Pos. & Rel Pos. & \#buckets & epoch & $AP$ & $AP_{50}$ & $AP_{75}$ & $AP_S$ & $AP_M$ & $AP_L$ \\
    \Xhline{2\arrayrulewidth}
    1~\cite{detr} & sinusoid & none & - & 150 & 39.5 & 60.3 & 41.4 & 17.5 & 43.0 & 59.1 \\ % official result (github)
    2 & none & none & - & 150 &  30.4{\textcolor{red}{\scriptsize{(-9.1)}}} & 52.5 & 30.2 & 9.4 & 31.2 & 50.5 \\
    3 & sinusoid & bias & $9 \times 9$ & 150 & 40.6{\textcolor{YellowGreen}{\scriptsize{(+1.1)}}} & 61.2 & 42.8 & 19.0 & 43.9 & 60.2 \\
    4 & none & contextual & $9 \times 9$ & 150 & 38.7{\textcolor{red}{\scriptsize{(-0.8)}}} & 60.1 & 40.4 & 18.2 & 41.8 & 56.7 \\ 
    5 & sinusoid & ctx clip & $9 \times 9$ & 150 & 40.4{\textcolor{YellowGreen}{\scriptsize{(+0.9)}}} & 60.9 & 42.4 & 19.1 & 43.7 & 59.8 \\
    6 & sinusoid & contextual & $9 \times 9$ & 150 & 40.8{\textcolor{YellowGreen}{\scriptsize{(+1.3)}}} & 61.5 & 42.5 & 18.5 & 44.4 & 60.5 \\
    7 & sinusoid & contextual & $15 \times 15$ & 150 & 40.8{\textcolor{YellowGreen}{\scriptsize{(+1.3)}}} & 61.7 & 42.6 & 18.5 & 44.2 & 61.2 \\
    \hline
    8~\cite{detr} & sinusoid & none & - & 300 & 40.6 & 61.6 & - & 19.9 & 44.3 & 60.2 \\ % official result (ablation part in paper)
    9 & sinusoid & contextual & $9 \times 9$ & 300 & 42.3{\textcolor{YellowGreen}{\scriptsize{(+1.7)}}} & 62.8 & 44.3 & 20.7 & 46.2 & 61.1 \\
    % 7 & sinusoid & none & - & 300 &  41.8 & 62.0 & 44.1 & 20.2 & 45.4 & 60.6 \\
    %10 & sinusoid & none & - & 500 & 42.0 & 62.4 & 44.2 & 20.5 & 45.8 & 61.1 \\ % official result (paper)
    \Xhline{2\arrayrulewidth}
\end{tabular}
    \caption {Component-wise analysis on DETR~\cite{detr}.}

    \label {table:det_comp}
    %\vspace{-3mm}
\end{table*}

\subsection{Comparison on Image Classification}
We compare our proposed methods with the state-of-the-art methods on image classification tasks. We select DeiT~\cite{deit} as the baseline. We adopt contextual Product shared-head method with a buckets number of 50. As shown in Tab. \ref{tab:sota_cls}, our method brings improvement on all three DeiT models. In particular, we improve the DeiT-Ti/DeiT-S/DeiT-B models by 1.5\%/1.0\%/0.6\% respectively, through adding 
relative position encoding only on keys. We show that the models could be further improved by adding the proposed relative position on both queries and values. When compared with other methods, ours achieve superior performance with less parameters and MACs.

\subsection{Comparison on Object Detection}

To verify the generality of our method, we further evaluate it on COCO 2017 detection dataset~\cite{coco}. We use the transformer-based detection model DETR~\cite{detr} as our baseline. We follow the same training and testing settings (including hyperparameters) as DETR~\cite{detr}, except injecting relative position encoding into all self-attention modules in the encoder. As shown in Tab.~\ref{table:det_comp} (\#1,6 and \#8,9), our method consistently improve the performance of DETR by 1.3mAP and 1.7mAP under 150 and 300 training epochs. 

In addition, we conduct ablation studies analyzing that the effects of position encoding on object detection task. Comparing \#1, \#2 and \#4 in Tab.~\ref{table:det_comp}, we give the conclusion that position encoding is crucial for DETR. We also show that absolute position embedding is better than relative position embedding in DETR, which is contrast to the observation in classification. We conjecture that DETR needs the prior of absolute position encoding to locate objects.

\begin{figure}[htbp]
\vspace{-4mm}

    \subfloat[][block 0]{
        \includegraphics[width=3.9cm]{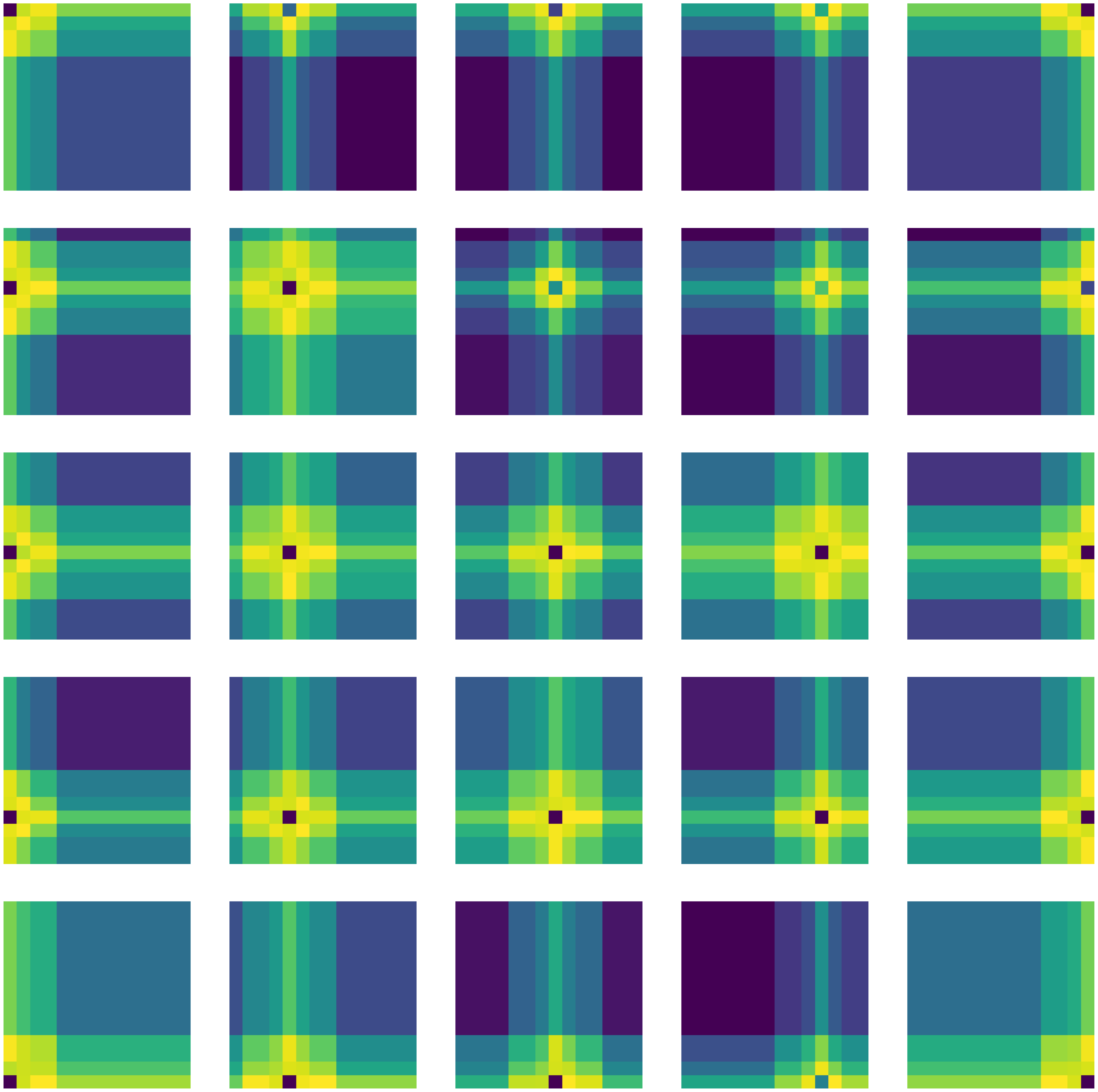}
    }
    \begin{minipage}[b]{1.0\textwidth}

    \hspace{0.8mm}
    \subfloat[][block 10]{
        \includegraphics[width=3.9cm]{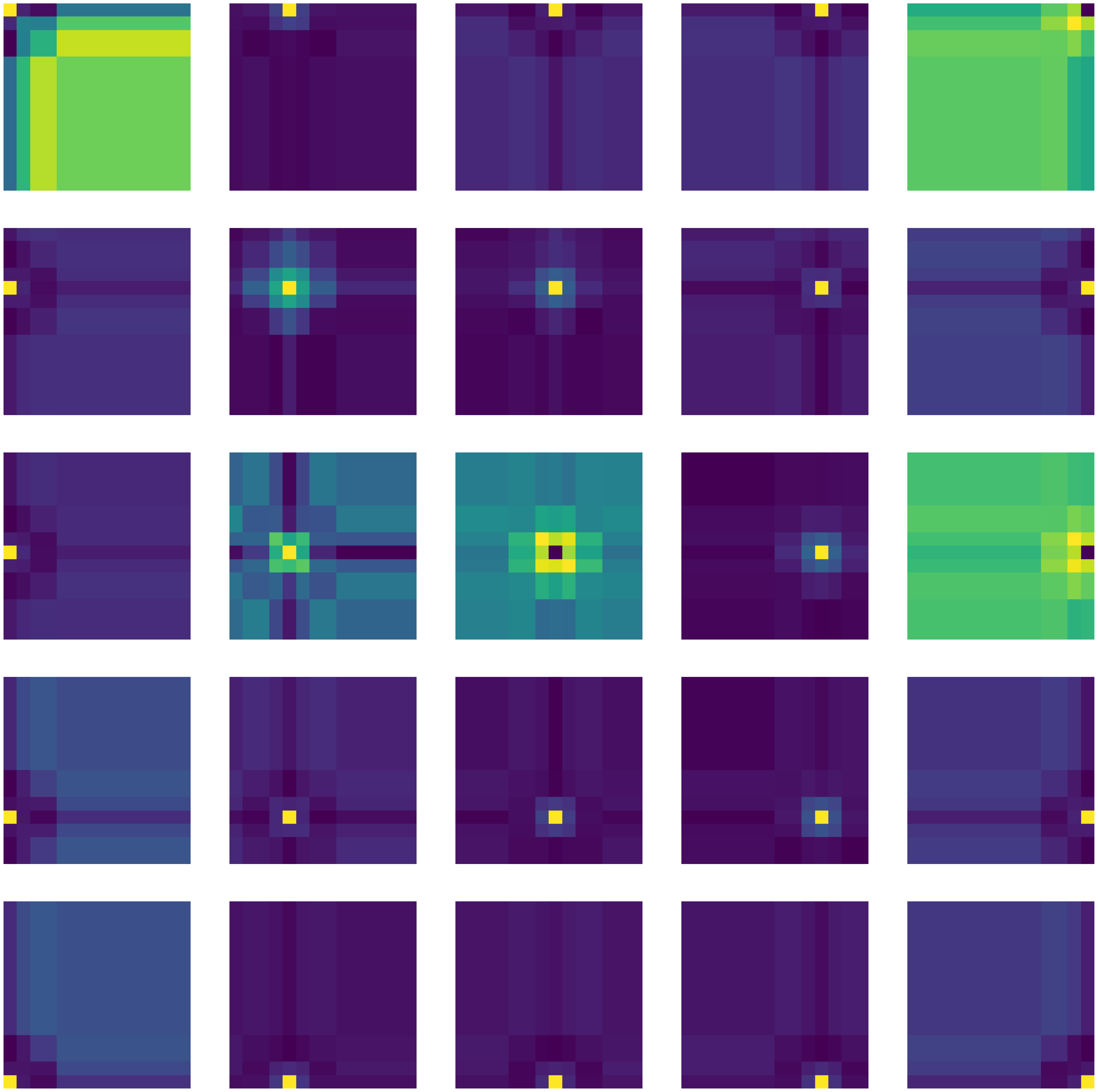}
    }
    \end{minipage}
    \caption {Visualization of relative position encoding (RPE) in contextual product method. We show the extra weights added to the attention by relative position encoding for different position. (a), (b) display the extra weights on attention for $5 \times 5$ reference patches uniformly sampled from $14 \times 14$ patches in block 0 and 10. }
    %\vspace{-3mm}

    \label {fig:vis_rpe}
\end{figure}

\subsection{Visualization}

We explore the underlying reason of relative position encoding in this subsection. We visualize the extra weights ${{b}}_{ij}$ (defined in Eq.~(\ref{eq:eindex})) added into the attention by relative position encoding for different positions in each block. %The model is trained on ImageNet with our methods with the number of 50 buckets. 
From Fig.~\ref{fig:vis_rpe}, we could see that relative position encoding makes the current patch focus more on its neighboring patches in block 0. However, when it turns to higher block, this phenomenon disappears. We conjecture this is because after passing through multiple layers, the model has already captured enough local information. The shallow layers in transformer are also global attentions, which pay attention to the whole image (consisting of small patches). It is different from CNN models in which shallow layers only capture local information. In theory, without RPEs (or other additional operations such as local windows). transformer does not explicitly capture locality. RPEs inject Conv-like inductive bias (including locality) into transformer, improving the model capability of capturing local patterns.

\section{Related Work}

\textbf{Transformer.} Transformer was originally introduced by Vaswani \emph{et al.}~\cite{Attention} for natural language processing, and recently extended to computer vision~\cite{ViT,deit,detr}.
%It is superior to capture long-range dependencies, and has been widely used in natural language processing \cite{bert, xlnet, T5, XL-transformer}. Recently, transformer is also extended to other fields, like image classification \cite{ViT, deit}, object detection \cite{detr, defor-detr}, semantic segmentation \cite{setr}, object tracking \cite{trackformer}, and point cloud classification \cite{pt, pct}. 
In this work we study vision transformers in image classification and object detection tasks, and select DeiT~\cite{deit} and DETR~\cite{detr} as our baseline models.
In ViT~\cite{ViT} and DeiT~\cite{deit} models, an image is split into multiple fixed-size patches. The embedded features of patches are added with absolute position encoding to fed in a standard transformer encoder. An extra trainable classification token is added into the sequence for classification. 
In DETR~\cite{detr}, a CNN backbone is used for feature extraction first. It outputs a feature map downsampled $32 \times$. Then it is flatten and fed to a transformer. The transformer outputs a certain number of bounding boxes. A learnable or sinusoid absolute position encoding is added in both transformer encoder and decoder. 

\textbf{Relative Position Encoding.} 
Relative position encoding is proposed firstly by Shaw \emph{et al.}~\cite{shaw}, where relative position encodings are added into keys and values. Dai \emph{et al.}~\cite{XL-transformer} proposed relative position encoding with the prior of the sinusoid matrix and more learnable parameters. Huang \emph{et al.}~\cite{better_rpe} proposed several 1D encoding variants. The effectiveness of relative position encoding has been verified in natural language processing.  There are also some works utilizing relative position encoding on 2D visual tasks. Ramachandran \emph{et al.}~\cite{sasa,bot} proposed 2D relative position encoding that computes and concatenates separate encodings of each dimension. Chu \emph{et al.}~\cite{meituan_rel} proposed position encoding generator, inserted between encoders. However, the efficacy of relative position encoding in visual transformer is still unclear, which is discussed and addressed in this work.

\section{Conclusions and Remarks}
In this paper, we review existing relative position encoding methods, and propose four methods dedicated to visual transformers. 
The abundant experiments show that our methods bring a clear improvement on both classification and detection tasks with negligible extra complexity. Our method could be easily plugged into the self-attention modules in vision models. In addition, we give comparison of different methods and analysis on relative position encoding with following conclusions.
1) Relative position encoding can be shared among different heads for parameter-saving. It is able to achieve comparable performance with the non-shared one in contextual mode.
2) Relative position encoding can replace absolute one in image classification task. However, absolute position encoding is necessary for object detection task, which needs to predict locations of objects.
3) Relative position encoding should consider the positional directivity, which is important to structured 2D images. 
4) Relative position encoding forces the shallow layers in transformers to pay more attention to local patches.

In future work, we plan to extend our method to other attention-based models and scenarios, such as high-resolution input tasks like semantic segmentation~\cite{setr}, and non-pixel input tasks like point cloud classification~\cite{pt, pct}.

%------------------------------------------------------------------------
\if 0
\section{Final copy}

You must include your signed IEEE copyright release form when you submit
your finished paper. We MUST have this form before your paper can be
published in the proceedings.
\fi

{\small
\bibliographystyle{ieee_fullname}
\bibliography{arxiv}
}

\clearpage

%%%%%%%%% ABSTRACT

% title
\twocolumn[
\begin{@twocolumnfalse}
   \vskip .375in
   \begin{center}
      {\Large \bf Rethinking and Improving Relative Position Encoding for Vision Transformer \\
------ Supplementary Material ------\par}
      % additional two empty lines at the end of the title
      %\vspace*{24pt}
      % additional small space at the end of the author name
      %\vskip .5em
      % additional empty line at the end of the title block
      \vspace*{12pt}
   \end{center}
\end{@twocolumnfalse}
]

\setcounter{section}{0}

This supplementary material presents additional details of Section $3.2$, $4.2$, $4.3$ and $4.4$. Besides, two extra experiments are added to demonstrate the effectiveness and generality of the proposed iRPE. We also provide comparisons on the inference time.

\begin{itemize}[noitemsep,topsep=0pt,parsep=0pt,partopsep=0pt]
    \item \textbf{Visualization of 2D relative position.} To provide an intuitive understanding, we visualize the proposed 2D relative position in Section $3.2$, including Euclidean, Quantization, Cross and Product methods.  
    \item \textbf{Weight initialization.} We elaborate the weight setting of the proposed relative position encoding methods, including the weight initialization and whether to equip with weight decay.
    \item \textbf{Computation complexity.} We provide a detailed explanation of why the computational costs are the same for shared and unshared relative position encodings across attention heads in Tab. \ref{table:buckets_sharing} of Section $4.2$.
    
    \item \textbf{Injecting previous RPE methods into DeiT.} We elaborate how to inject previous relative position encoding methods into DeiT~\cite{deit} in Tab. \ref{tab:sota_cls} of Section $4.3$.

    \item \textbf{Training and test settings of DETR.} We provide the details of training and test settings of DETR~\cite{detr} in Section $4.4$.
    
    \item \textbf{The effectiveness on other vision transformers.} We show the effectiveness of the proposed iRPE on the recent Swin transformer \cite{swin}.
 
    \item \textbf{Transfer learning on fine-grained datasets.} {To verify the generalizability, we evaluate our models on fine-grained datasets, including Stanford Cars and CUB200\_2011 datasets.}

    \item \textbf{Inference performance.}
    {We compare the proposed iRPE with previous methods in terms of inference time and memory cost.}

\end{itemize}

\section{Visualization of 2D Relative Position}

\begin{figure}[t]

    \vspace{-4mm}
    \begin{minipage}[b]{1.0\textwidth}
    %\centering
    \subfloat[][top-left]{
        \includegraphics[width=4.1cm, trim=40 0 80 0, clip]{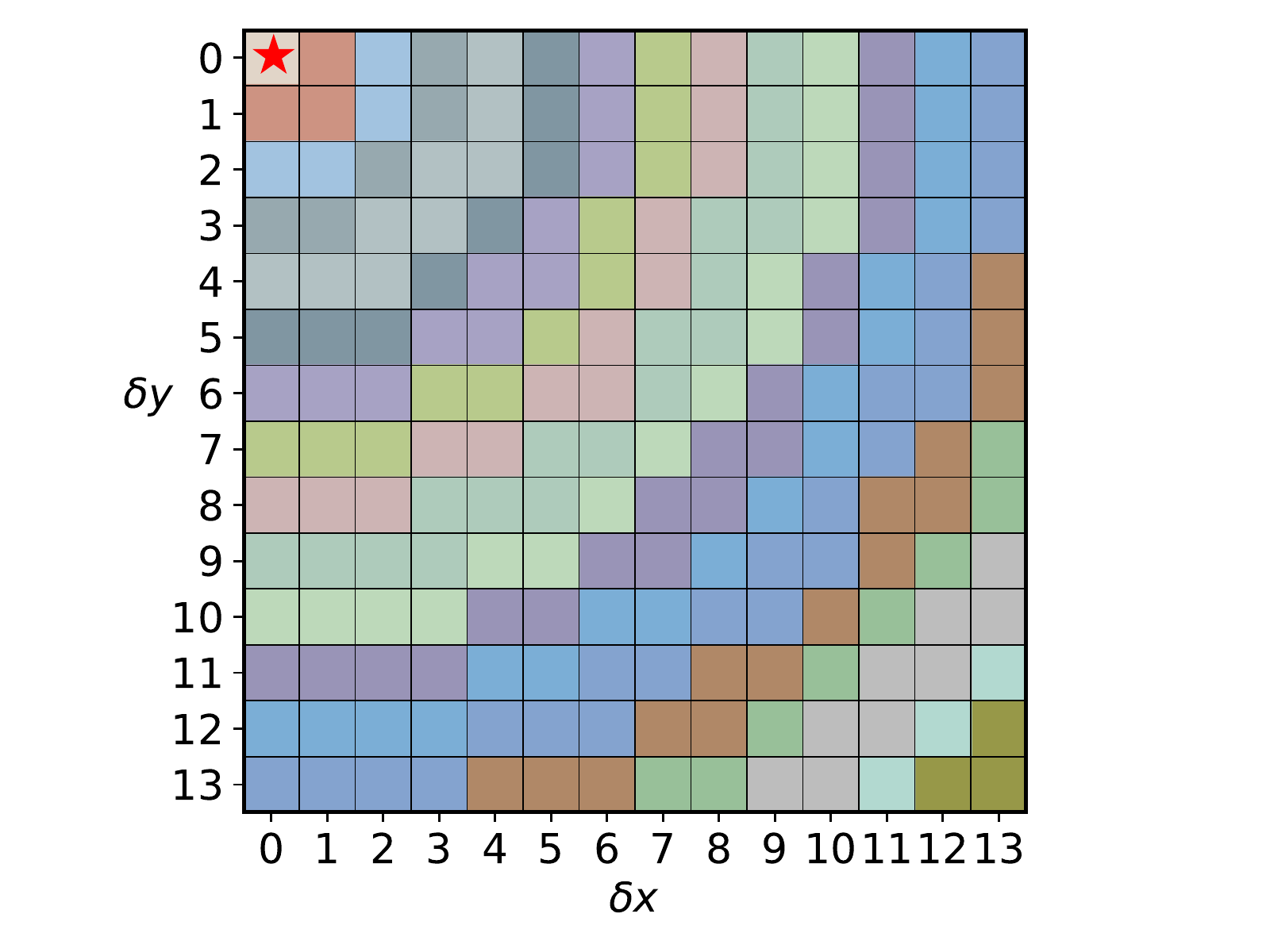}
    }
    %\centering
    \subfloat[][center]{
        \label {fig:vis_euc_rpe_center}
        \includegraphics[width=4.1cm, trim=40 0 80 0, clip]{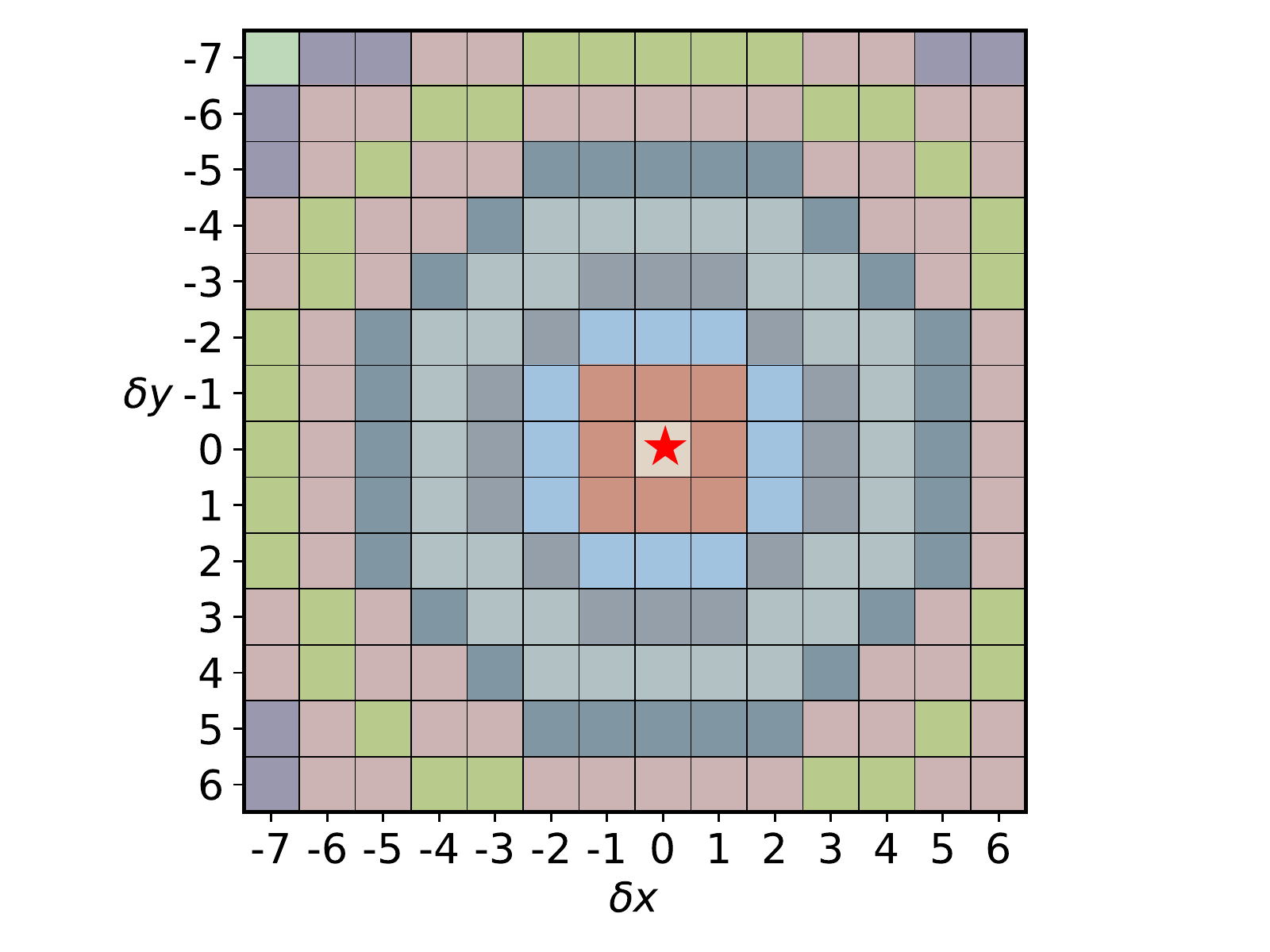}
    }
    \end{minipage}

    \vspace{-3mm}
    \caption {Visualization of Euclidean method. The red star \textcolor{red}{$\bigstar$} presents the reference position. Different color means different bucket. The relative positions with the same color share the same  encoding. }

    \label {fig:vis_euc_rpe}
    %\vspace{-4mm}
\end{figure}

\begin{figure}[t]
    \vspace{-3mm}
    \begin{minipage}[b]{1.0\textwidth}
    %\centering
    \subfloat[][top-left]{
        \includegraphics[width=4.1cm, trim=40 0 80 0, clip]{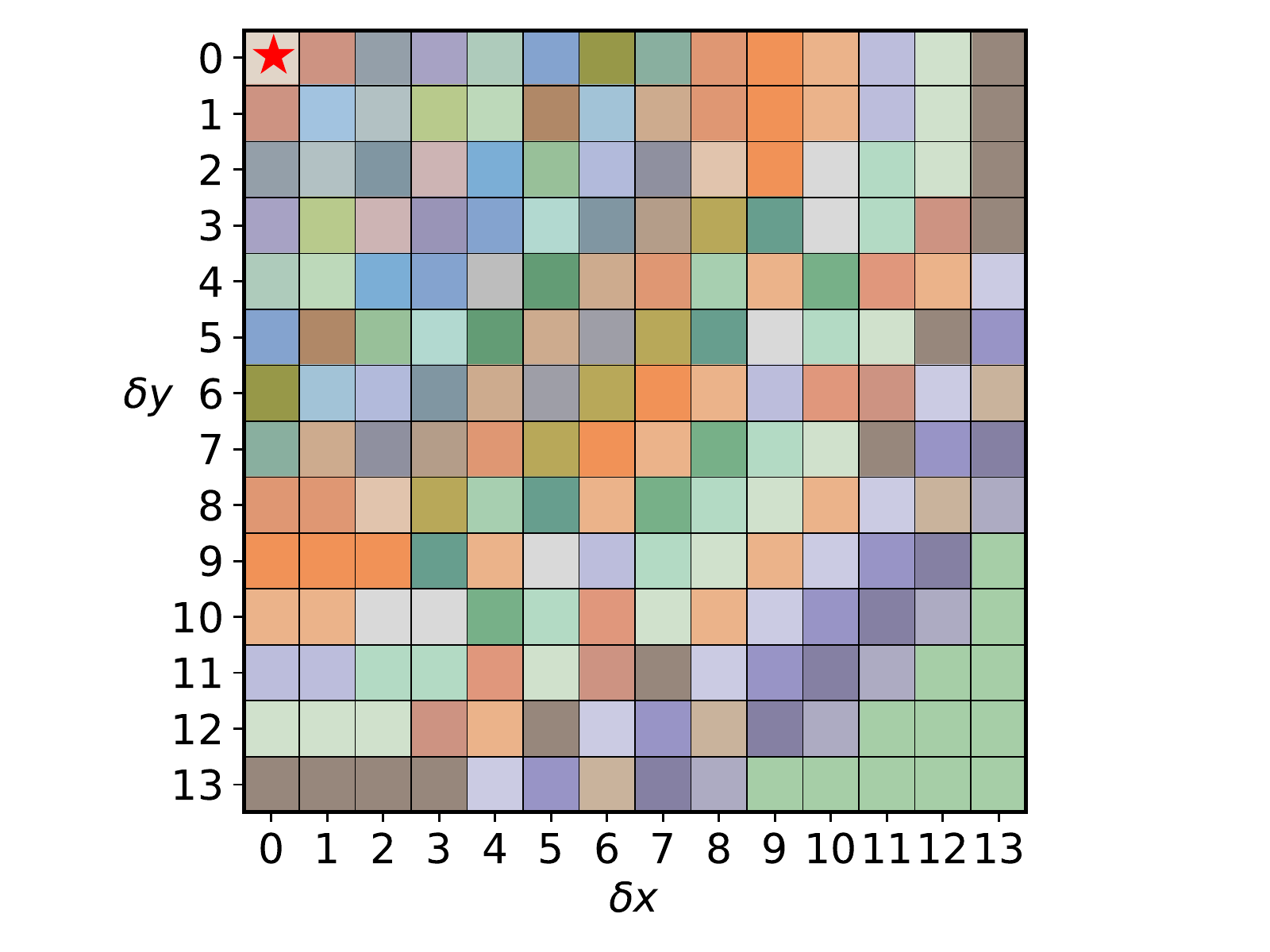}
    }
    %\centering
    \subfloat[][center]{
        \label{fig:vis_quant_rpe_center}
        \includegraphics[width=4.1cm, trim=40 0 80 0, clip]{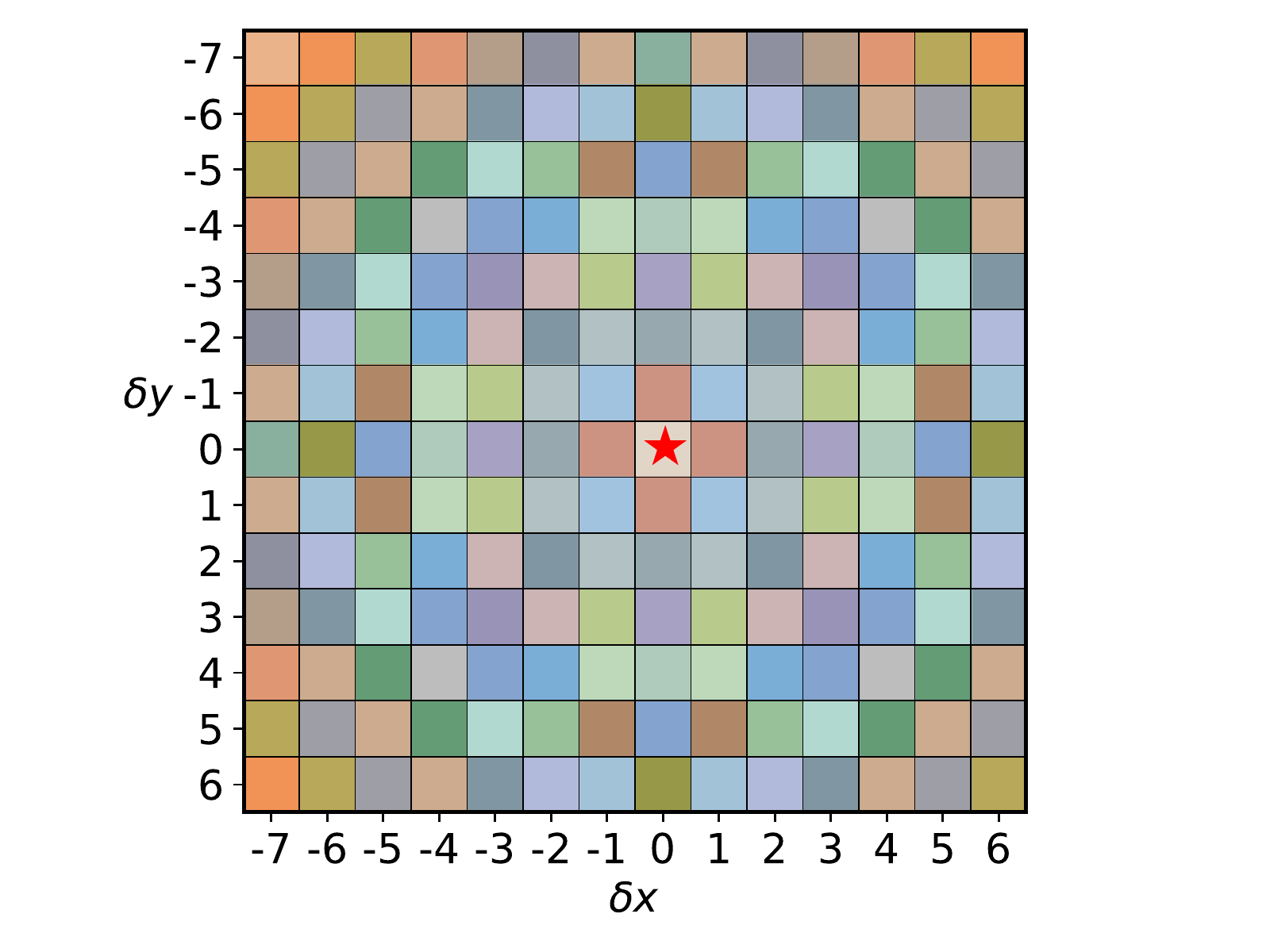}
    }
    \end{minipage}
    \vspace{-3mm}
    \caption {Visualization of Quantization method. The red star \textcolor{red}{$\bigstar$} presents the reference position. Different color means different bucket. The relative positions with the same color share the same encoding. }

    \label {fig:vis_quant_rpe}
    %\vspace{-4mm}
\end{figure}

\begin{figure}[t]
    \vspace{-4mm}
    \begin{minipage}[b]{1.0\textwidth}
    %\centering
    \subfloat[top-left (horizontal)]{
        \includegraphics[width=4.1cm, trim=40 0 80 0, clip]{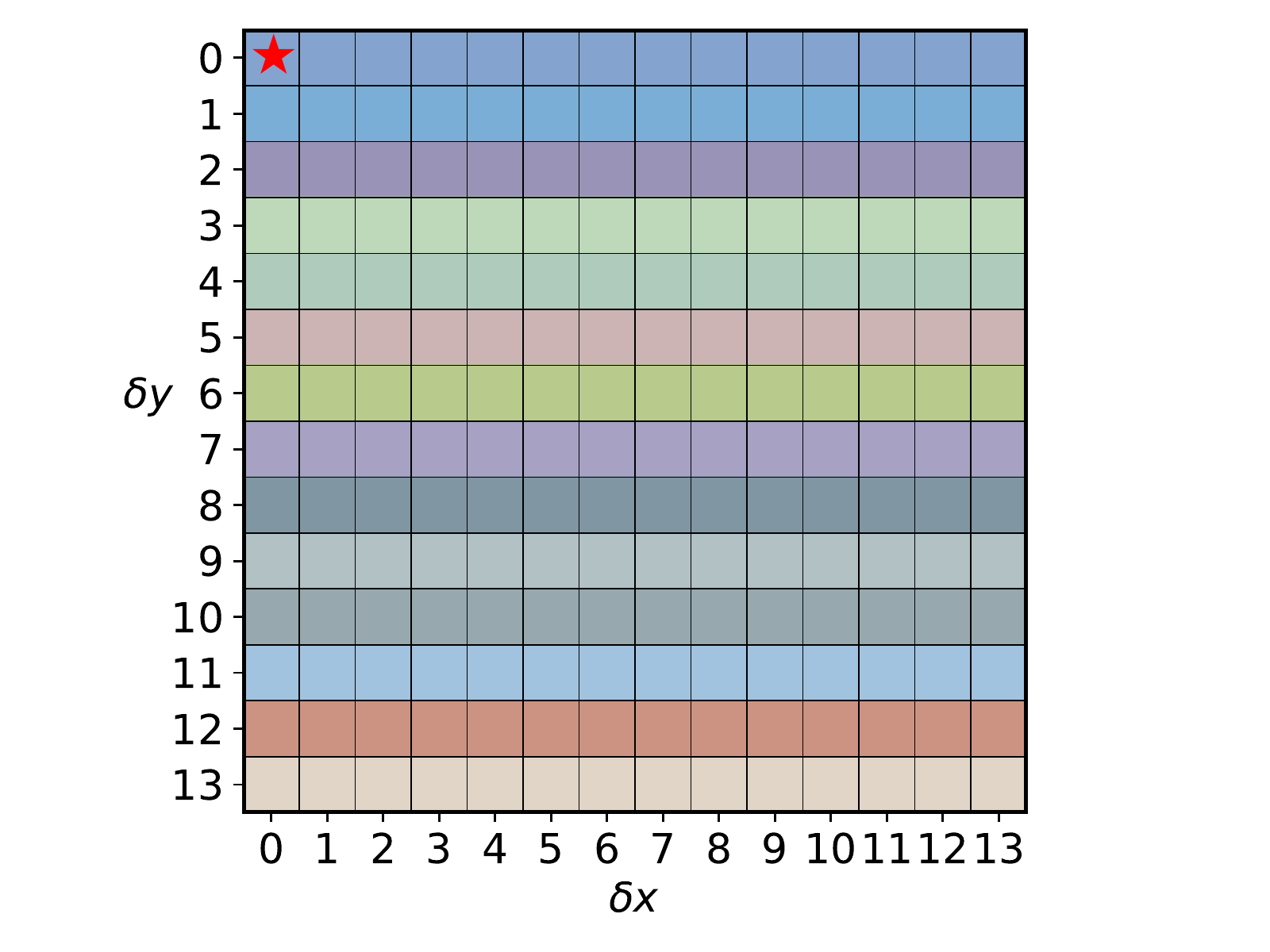}
    }
    %\centering
    \subfloat[top-left (vertical)]{
        \includegraphics[width=4.1cm, trim=40 0 80 0, clip]{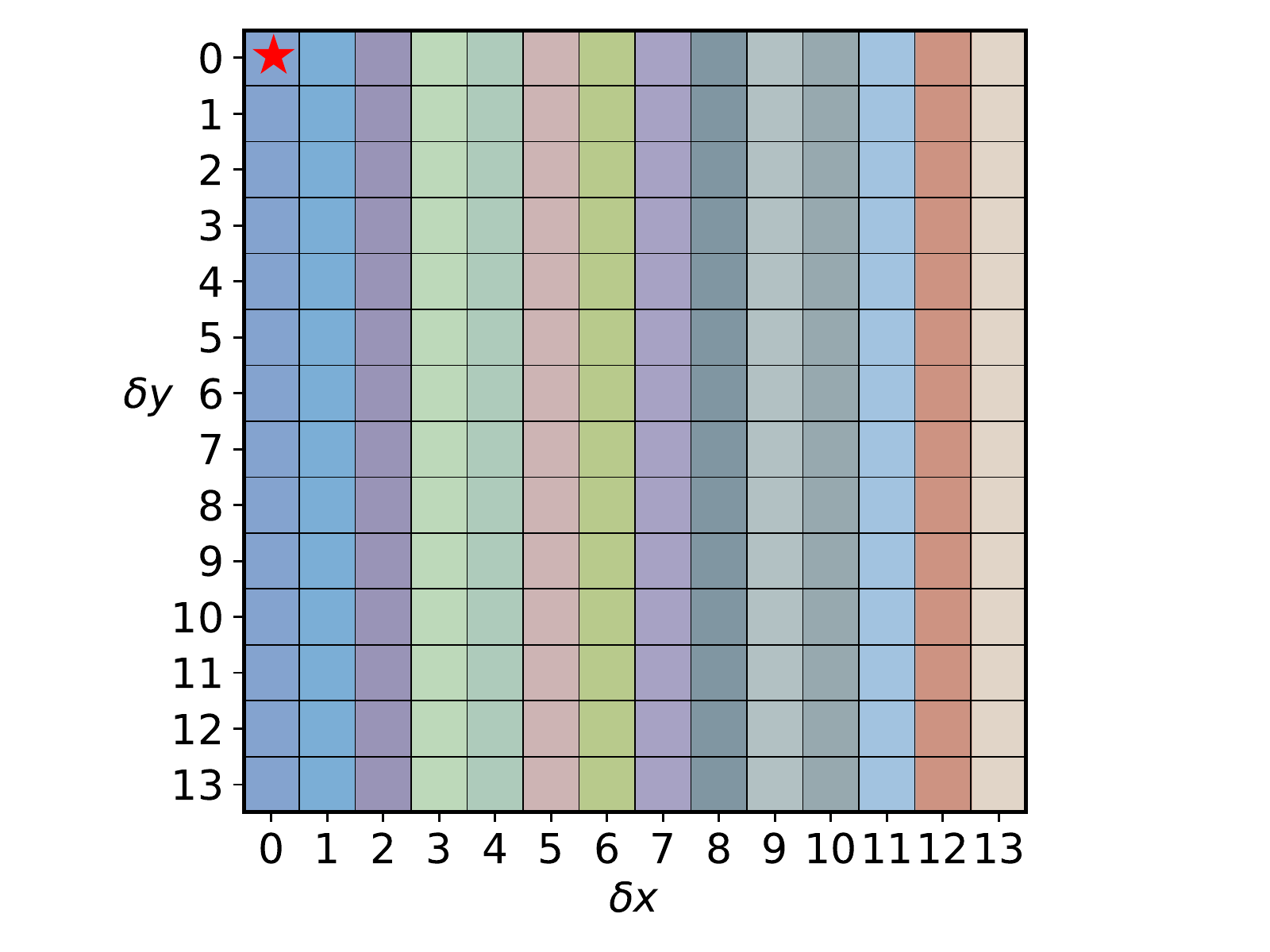}
    }
    \end{minipage}
    
    \vspace{-2mm}
    \begin{minipage}[b]{1.0\textwidth}
    %\centering
    \subfloat[center (horizontal)]{
        \label{fig:vis_cross_rpe_center_h}
        \includegraphics[width=4.1cm, trim=40 0 80 0, clip]{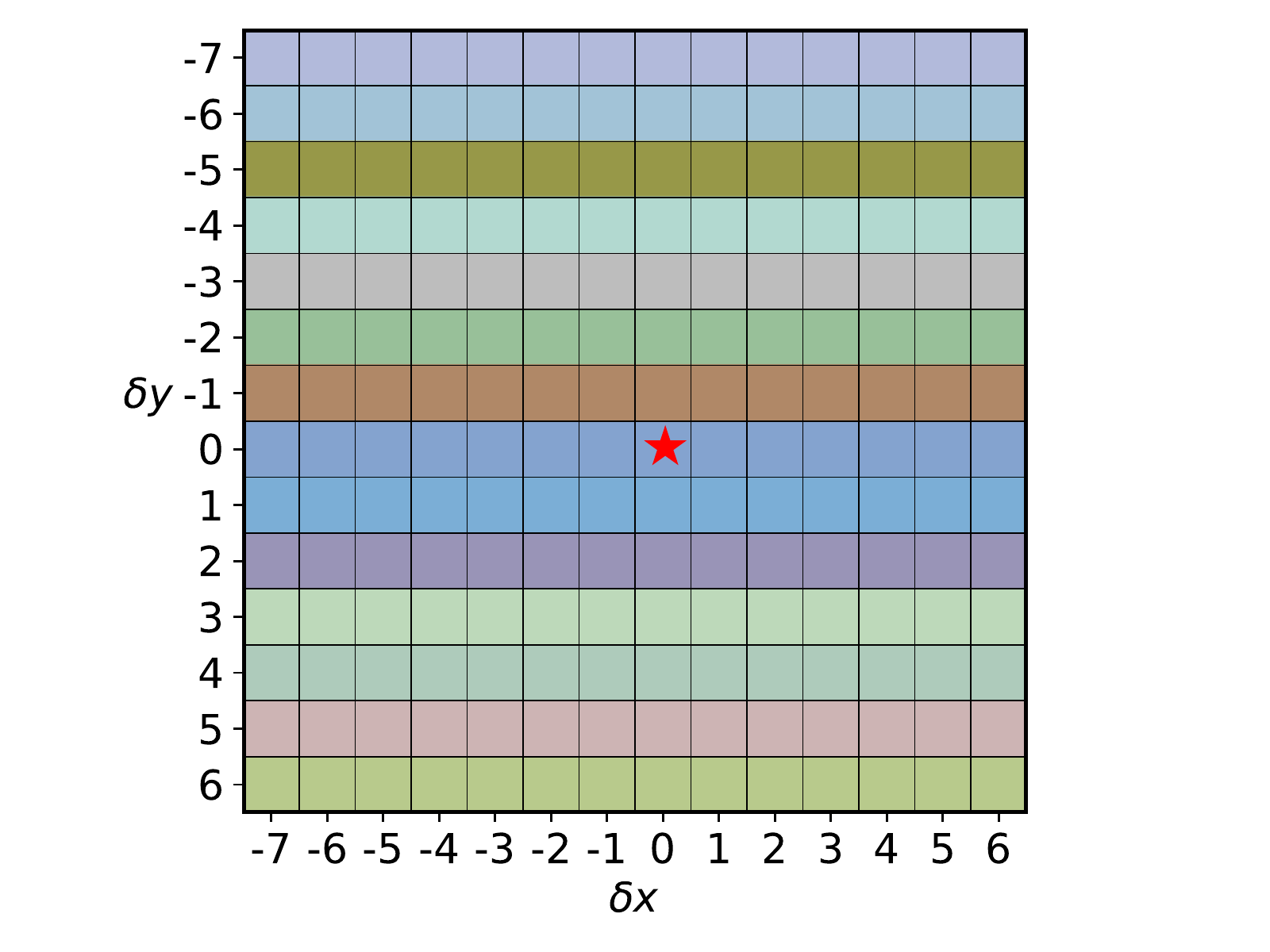}
    }
    %\centering
    \subfloat[center (vertical)]{
        \label{fig:vis_cross_rpe_center_v}
        \includegraphics[width=4.1cm, trim=40 0 80 0, clip]{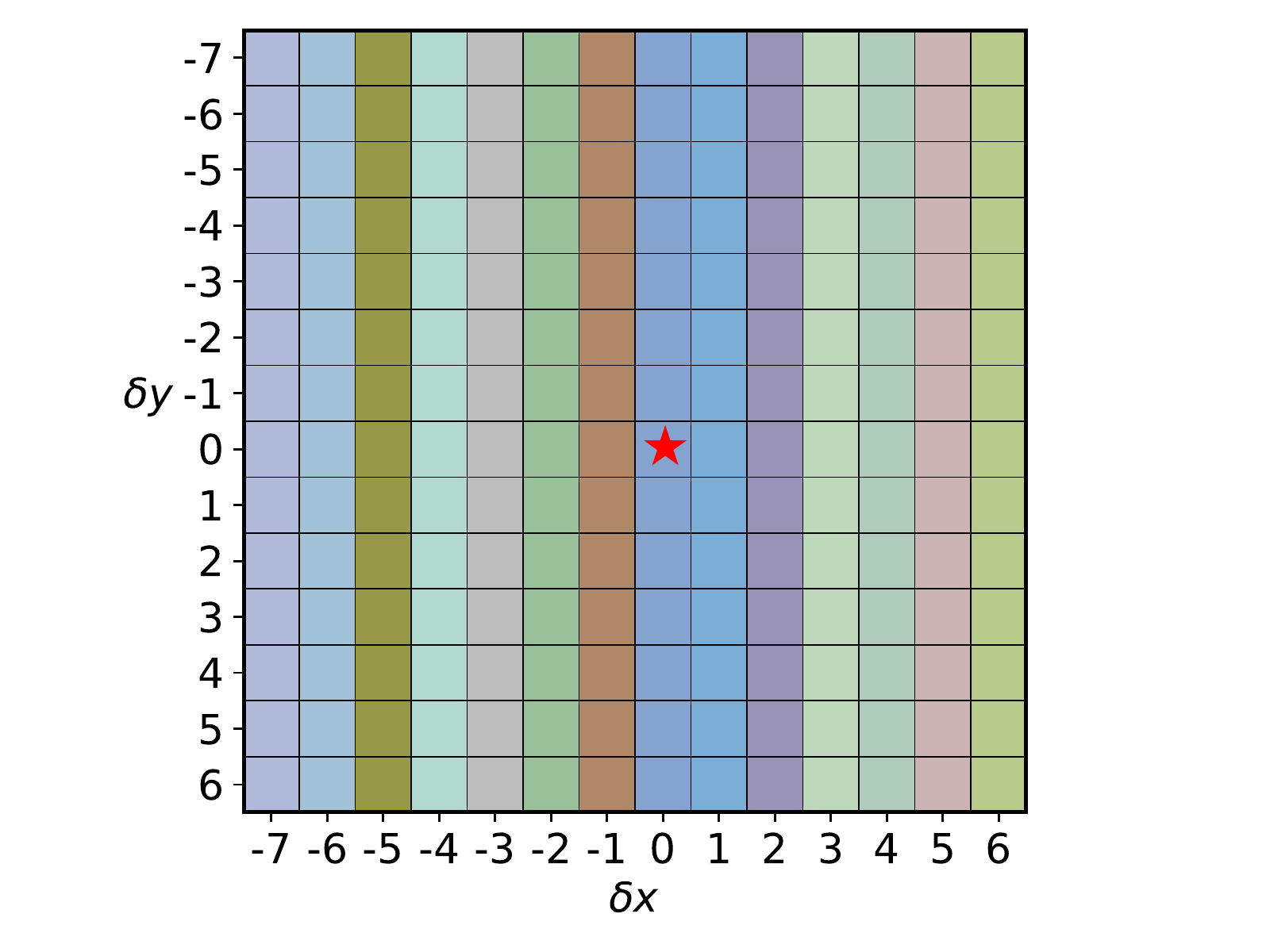}
    }
    \end{minipage}
    \vspace{-3mm}
    
    \caption {Visualization of Cross method. The red star \textcolor{red}{$\bigstar$} presents the reference position. Different color means different bucket. The relative positions with the same color share the same encoding. }
    \label {fig:vis_cross_rpe}
    %\vspace{-4mm}
\end{figure}

\begin{figure}[t]

    \vspace{-4mm}
    \begin{minipage}[b]{1.0\textwidth}
    %\centering
    \subfloat[][top-left]{
        \includegraphics[width=4.1cm, trim=40 0 80 0, clip]{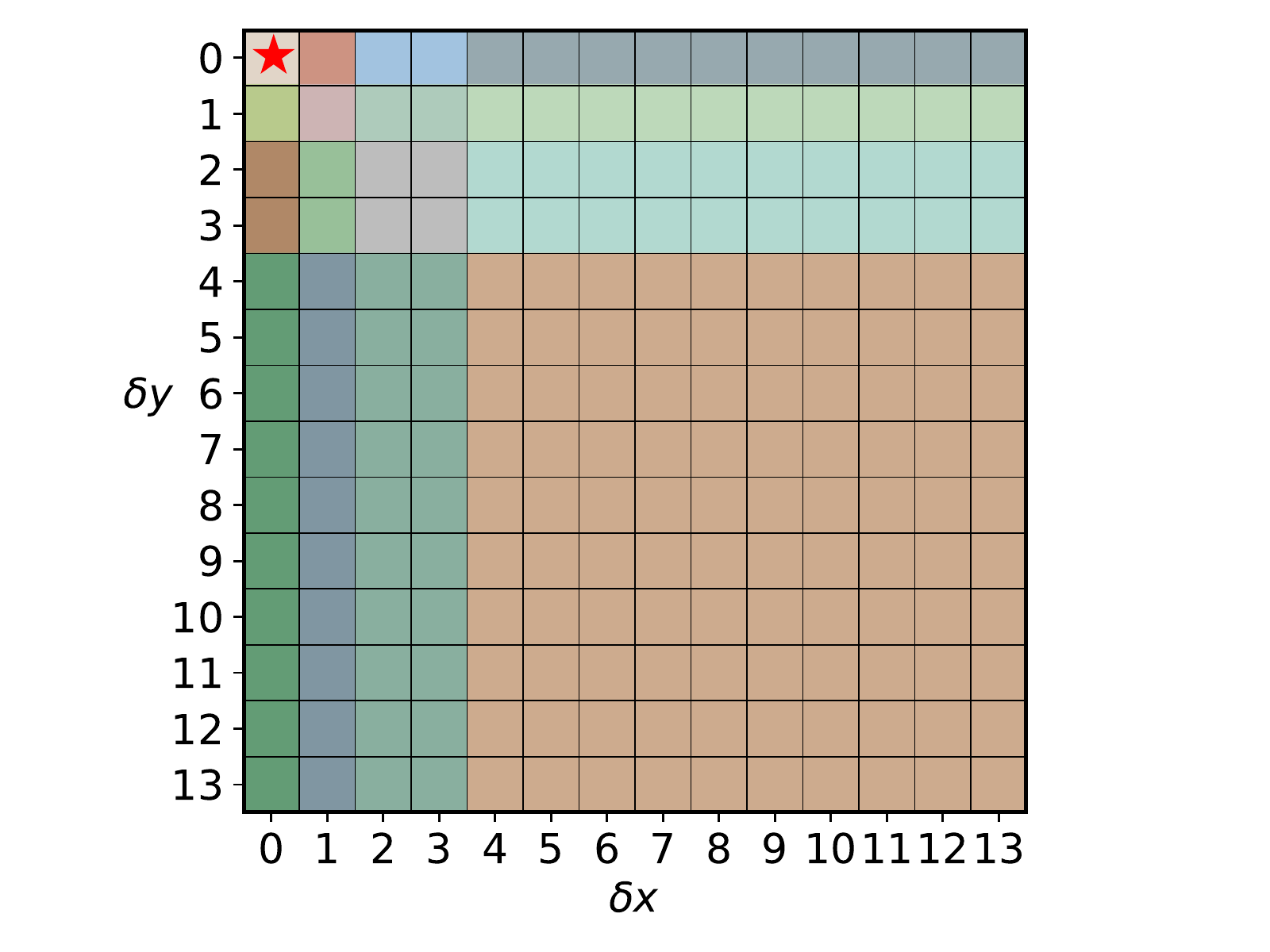}
    }
    \subfloat[][center]{
        \label{fig:vis_product_rpe_center}
        \includegraphics[width=4.1cm, trim=40 0 80 0, clip]{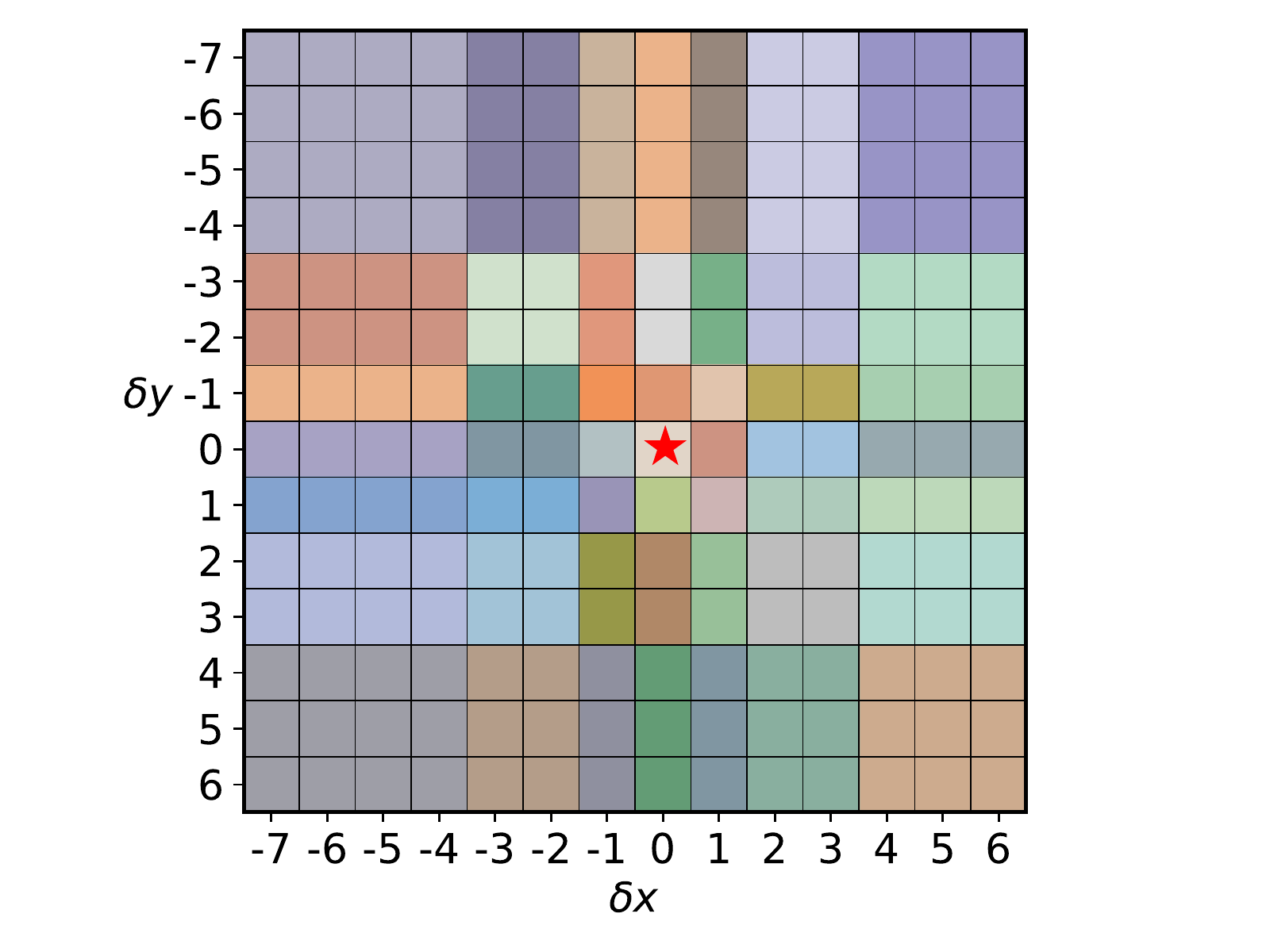}
    }
    \end{minipage}

    \vspace{-3mm}
    \caption {Visualization of Product method. The red star \textcolor{red}{$\bigstar$} presents the reference position. Different color means different bucket. The relative positions with the same color share the same encoding. }

    \label {fig:vis_product_rpe}
    %\vspace{-4mm}
\end{figure}

We visualize the proposed four relative position methods, \emph{i.e.}, Euclidean, Quantization, Cross and Product, and present their difference.
In DeiT~\cite{deit}, an image is split into $14 \times 14$ non-overlapping patches, so the number of tokens is $14 \times 14$ (except for the classification token).  Therefore, in theory, each token has $14 \times 14$ relative positions. For visualization, we select the top-left position $(0, 0)$ and the center position $(7, 7)$ as the reference positions (presented by a red star \textcolor{red}{$\bigstar$} in the following figures), and then compute the relative offsets $\delta x = x_i - x_j$ and $\delta y = y_i - y_j$ between the reference position and the remaining $14 \times 14 -1$ positions. Let $(\delta x, \delta y)$ denote a 2D relative position. We plot the map of the relative encoding ${\bf{r}}_{ij}$, defined in Eq.~(\ref{eq:euc_r}), Eq.~(\ref{eq:method_quant}), Eq.~(\ref{eq:method_cross}) and Eq.~(\ref{eq:method_product}), where $i$ is the reference position and $j$ is one of the $14 \times 14$ positions. Notice that ${\bf{r}}_{ij}$ is either a learnable scalar in bias mode or a vector in contextual mode. Multiple ${\bf{r}}_{ij}$ may share an identical bucket, which is presented by the same color in Fig.~\ref{fig:vis_euc_rpe} - \ref{fig:vis_product_rpe}. Different bucket is presented by different color.

\textit{Euclidean method.} Fig.~\ref{fig:vis_euc_rpe} shows Euclidean method. It is an undirected method, since the relative position encodings only depend on relative Euclidean distance. For example, in Fig.~\ref{fig:vis_euc_rpe_center} since the relative positions $(-1, 0)$ and $(1, 0)$ have the same relative Euclidean distance of 1, they are mapped into the same bucket (the grids with \textcolor{orange}{orange} color).

\textit{Quantization method.} Fig.~\ref{fig:vis_quant_rpe} presents Quantization method, another undirected method. It is an improved version of Euclidean method, and addresses the problem that the close two neighbors with different relative distances might be mapped into the same bucket (\emph{e.g.}, he relative position $(1, 0)$ and $(1, 1)$ are both mapped into the same bucket in Euclidean method). Besides, the number of buckets in Quantization method is larger than that in Euclidean method. The reason is that Quantization method quantize Euclidean distance from a set of real numbers \{0, 1, 1.41, 2, 2.24, ...\} to a set of integers \{0, 1, 2, 3, 4, ...\}, increasing the number of buckets for adjacent positions. %increasing the distance between adjacent positions, so the number of buckets can be larger.

\textit{Cross method.} Fig.~\ref{fig:vis_cross_rpe} shows Cross method. It is a directed method, in which the relative position encoding depends on relative distances and relative directions simultaneously. It computes the encodings on horizontal and vertical directions separately, then summarizes them. The same offsets along $x$-axis (or $y$-axis) direction share the same horizontal (or vertical) encoding. For example, the two relative positions $(-1, 0)$ and $(1, 0)$ share the same encoding on horizontal in Fig.~\ref{fig:vis_cross_rpe_center_h}, but not on vertical in Fig.~\ref{fig:vis_cross_rpe_center_v}.

\textit{Product method.} Fig.~\ref{fig:vis_product_rpe} shows Product method, which is also a directed method. Unlike Cross method, Product method does not share the same encoding even if the offsets are the same along x-axis or y-axis direction. For example, in Fig.~\ref{fig:vis_product_rpe_center}, the two relative positions $(-1, 0)$ and $(1, 0)$ have independent encodings. Moreover, it is more efficient than Cross method, since there is no extra addition operation in Eq.~(\ref{eq:method_cross}).

\section{Weight Initialization}
The relative position weight ${\bf{r}}_{ij}$ in Eq.~(\ref{eq:abs_bucket}) and Eq.~(\ref{eq:ctx_bucket}) is initialized with zero. We found that there is no difference between zero and normal-distribution initialization. Besides, we do not impose weight decay on the weight of relative position encodings, because its effects on the final performance is negligible. 

\section{Computation Complexity}
As shown in Tab. \ref{table:buckets_sharing} (in the main manuscript), the computational costs MACs of shared and unshared relative position encodings across attention heads are the same. Here, we provide the detailed explanation.
Let $h, n, d, k$ denote the number of heads, the length of a sequence, the number of channels and the number of buckets, respectively. For bias mode, in Eq.~(\ref{eq:eindex}), the broadcast addition on the dot-product attention $({\bf{x}}_i{\bf{W}}^Q)({\bf{x}}_j{\bf{W}}^K)^T$ with the shape of $h \times n \times n$ and the encoding $b_{ij}$ with the shape of $n \times n$ in shared scheme or $h \times n \times n$ in unshared scheme takes the computational cost of $\mathcal{O}(hn^2)$. For contextual mode, in Eq.~(\ref{eq:imp_zit}), the broadcast multiplication on the input embedding ${\bf{x}}_i{\bf{W}}$ with the shape $h \times n \times d$ and the relative position weight $\bf{p}$ with the shape of $d \times k$ in shared scheme or $h \times d \times k$ in unshared scheme takes the computational cost of $\mathcal{O}(hndk)$. Due to the broadcast operations, the computational cost of shared and unshared schemes is the same.

\section{Injecting Previous RPE Methods into DeiT}
In the Tab.~\ref{tab:sota_cls}, in order to compare with previous 1D relative position encoding methods, we utilize our Product method (defined in Sec.~$3.2$ in the main manuscript) to adapt 1D encoding methods for 2D images. We replace the piecewise function $g(x)$ with the clip function $h(x)$, which is matched with previous methods. The encoding weight is shared across attention heads. DeiT-S(Shaw's), DeiT-S(Trans.-XL's), DeiT-S(Huang's) are DeiT-S~\cite{deit} models with Shaw's relative position encoding~\cite{shaw}, relative position encoding in Transformer-XL~\cite{XL-transformer} and Huang's relative position encoding~\cite{better_rpe}, respectively. Besides, the 2D relative position encoding in SASA~\cite{sasa} is equipped on DeiT-S~\cite{deit} directly.

\begin{figure}[t]
\vspace{-4.4mm}
    \centerline{\includegraphics[width=9.0cm, height=4.0cm]{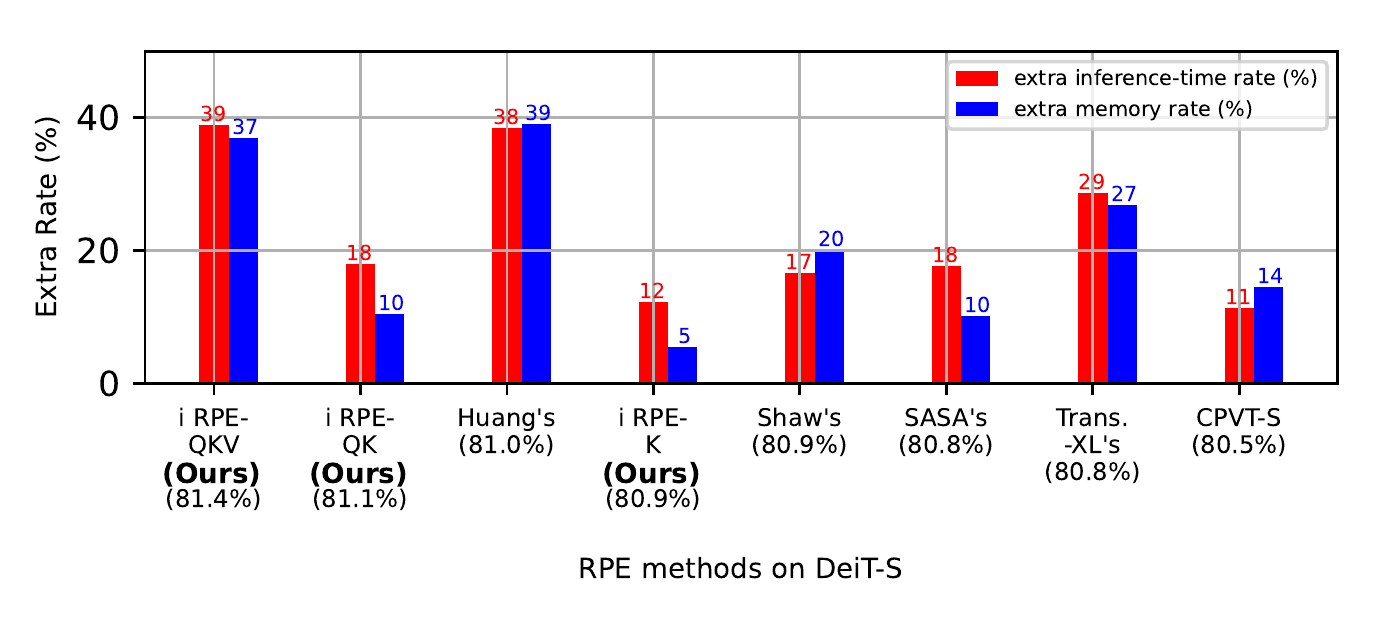}}
    \vspace{-6mm}
    \caption {{The extra cost brought by RPEs. The reference model is DeiT-S~\cite{deit} without RPE, taking 1,096 images/s and 8,930~Mb memory.}}
    \label {fig:cost}
    %\vspace{-3mm}
\end{figure}

\section{Training and Test Settings of DETR}
We follow the same training protocol and hyperparameter configurations as the original DETR \cite{detr}.
The backbone model of DETR~\cite{detr} is ResNet-50~\cite{resnet}, pretrained on ImageNet~\cite{imagenet}, and the BatchNorm layers are frozen during training. All transformer blocks are initialized with Xavier initialization~\cite{xavier}. The image is cropped such that the shortest side is at least 480 and at most 800 pixels while the longest at most 1333. When training, random horizontal flipping and random cropping are utilized. The initial learning rates of transformer and backbone are $10^{-4}$ and $10^{-5}$, respectively. Learning rates are divided by 10 in the last 50 epochs in 150 epochs schedule, and the last 100 epochs in 300 epochs schedule. The optimizer is AdamW~\cite{adamw} with weight decay of $10^{-4}$ and a mini-batch size of 16. The number of queries is 100. We train the models for 150 epochs and 300 epochs.

\section{The Effectiveness on Other Vision Transformers}
We further verify the effectiveness of the proposed iRPE on the recent Swin transformer~\cite{swin}. Specifically, the original Swin-T model without RPE obtains a top-1 accuracy of 80.5\% (Tab.~4 in Swin transformer~\cite{swin}), while using RPE bias mode gets +0.8\% improvements. Our contextual RPE on QKV can further improve Swin-T to 81.9\% on ImageNet.

\section{Transfer Learning on Fine-grained Datasets}
We finetune the pretrained models on Stanford Cars and CUB200\_2011 datasets using the resolution 224x224 and 300 epochs.
DeiT-B~\cite{deit} with iRPE on keys obtains a top-1 accuracy of 93.4\% and 84.9\% on the two datasets respectively, outperforming the original DeiT-B (92.1\% and 83.4\%) by 1.3\% and 1.5\% points.

\section{Inference Performance}
The inference runtime and memory cost are reported in Fig.~\ref{fig:cost}, tested on Nvidia V100 GPU with a batch size of 128. We can see that our iRPE on keys is more effective.

\end{document}